\documentclass{rob}

\usepackage{graphicx}
\usepackage{amsmath,amssymb} 
\usepackage{color}

\usepackage{times}
\usepackage{xcolor}
\usepackage{soul}
\usepackage[utf8]{inputenc}
\usepackage[small]{caption}

\usepackage{graphics,graphicx} 
\usepackage{float}
\usepackage{amsmath} 
\usepackage{amssymb}  
\usepackage{algorithmic}
\usepackage{algorithm}
\usepackage{verbatim}
\usepackage[normal]{subfigure}
\usepackage{url}

\usepackage{graphics} 
\usepackage{times} 
\usepackage{amsmath} 
\usepackage{amssymb,mathrsfs,wasysym}  
\usepackage{array}
\usepackage{url}

\usepackage{caption}

\graphicspath{ {./pic/} {./fig/}}

\title[Best Grasping Handle in a Clutter]{\LARGE \bf Optimized edge-based grasping method for a cluttered environment}

\author{Olyvia Kundu and Swagat Kumar}

\begin{document}

\maketitle

\begin{abs}
This paper looks into the problem of grasping region localization along with suitable pose from a cluttered environment without any a priori knowledge of the object geometry. This end-to-end method detects the handles from a single frame of input sensor. The pipeline starts with the creation of multiple surface segments to detect the required gap in the first stage, and eventually helps in detecting boundary lines. Our novelty lies in the fact that we have merged color based edge and depth edge in order to get more reliable boundary points through which a pair of boundary line is fitted.  Also this information is used to validate the handle by measuring the angle between the boundary lines and also by checking for amy potential occlusion. In addition, we also proposed an optimizing cost function based method to choose the best handle from a set of valid handles. The method proposed is tested on real-life datasets and is found to out form state of the art methods in terms of precision.     
\end{abs}

\section{INTRODUCTION} \label{sec:intro}
Human like manipulations in robots is a challenging research work which has attracted the research community a lot. Robots which can not only perceive the environment but also can manipulate the environment will be more helpful in carrying out many monotonous tasks and relieve the human beings for more creative pursuit. Among the manipulation abilities of a robot, the grasping ability remains a difficult problem that has attracted attention in the research community in the recent past. The methods existing in the literature that deals with grasping abilities can be classified in two categories – (1) relying on the availability of accurate geometric information about the object (or a CAD model) or multiple frames of objects to train a deep learning network (2) computing grasp affordances and pose directly from RGBD point cloud without any a-priori knowledge about the shape or heavily trained network.

We have followed the second approach where the grasping handles are computed from a single view of the point cloud rather than stitching multiple input frames together obtained from different views of the points. Although incomplete information makes the task much more difficult, but this type of scenario makes the system more flexible for real life application. The CAD model based technique was famous in grasping society due to good performance in various work \cite{miller2004graspit}\cite{klank2009real}. But now-a-days, its use is also going down as with the increase of object varieties it is quite impossible to get precise geometric information (CAD model). Also, the increasing use of deep learning techniques in this domain diminishes the use of CAD model based matching for grasping location extraction. Some work\cite{vezzani2017grasping} model the grasping location using superquadratic function. Lot of work applies deep learning technique \cite{lenz2015deepgrasp} \cite{johns2016deep}
\cite{pinto2016supersizing} \cite{Schwarz:7139363} to produce good results, but the downside is deep learning techniques \cite{mahler2017dex} also require huge amount gathering of data which is time-consuming and also off-line training processes which limit their applications to a vast number of real world problems. 

This paper deals with finding graspable affordances for a given object in a cluttered environment. This is known as the problem of grasp pose detection (GPD). Find a graspable affordance in a confined and cluttered workspace is more challenging as compared to detecting handles in isolated object scenario described in \cite{JainICRA16}\cite{tableTopGrasping}\cite{fischinger2012shape}. In this paper, we investigate the problem of grasp pose detection (GPD) and find suitable graspable affordance for a given object in a cluttered environment. The grasping problem is difficult in this case as the robot has to detect a handle within a confined and cluttered workspace. This is particularly more challenging problem as most of existing work focus on detecting handles in isolated object scenario as in \cite{JainICRA16}\cite{tableTopGrasping}\cite{fischinger2012shape}. The handles to be detected from a 3D point cloud obtained using a depth sensor from a single viewpoint.  Our method is motivated by platt’s work in \cite{Pas2013LocalizingGA}\cite{Pas2015UsingGT}, where they primarily rely on surface curvature to separate out potential grasping regions. their approach has a 
huge advantage that they do not need any information about the object beforehand. The main drawback of their work is that they randomly draw points from the workspace and hence, they might miss some valid handle just because no points are drawn from the
area. The region associated with these points is a sphere of fixed radius. If the dataset contains objects with different size, then [13][11] can only detect handles that are close to the specified radius. Also, they do not explore surface discontinuities and the results for rectangular shaped objects are not that promising. 

In this work, we overcome above described limitations by defining object boundaries with the help of surface discontinuities and surface normal. This information is further utilized to automatically determine the scale of the grasping region and in later stage for identifying a pair of boundary lines. This information is used by region growing algorithm that groups the points which have similar type of characteristics. These regions are like superpixels, where the feature of the region is represented by the whole group of points altogether. In the next step, the grasping algorithm is again divided into three parts. In the initial part, a grasp hypothesis generation step finds a required amount of gap around the handle. To validate each and every handle we propose a novel method that merges edges from color image and depth image to get a pair of boundary lines. These boundary lines serve two purposes. First,- the angle between these two lines play a crucial part to validate the hypothesis. The interaction between the gripper plate and the object is expected to happen on these boundary lines. As, in this work, a two finger parallel gripper will be used to pick the handle, where the two end plate is parallel to each other, then the boundary line also has be parallel to each other. Second,-these boundary lines also help to form a filter which easily detects any obstacle that lies on the path of the handle.

A popular idea in the domain of grasping is to extract visual features for 2D images of the objects. The authors of \cite{saxena2008robotic} applied some predefined edge and texture filters to capture some features and trained a probabilistic model to predict a grasping point. The paper \cite{6094932} also found out 2D edges and group those edges into contours. If two contours have similar characteristic, then they used edge-based grasp. This same process is applied for small 3D patches and then extended to find boundary surflings. Our work differentiates from the methodology of \cite{6094932} by merging both color and depth edges in order to get meaningful points and fit boundary lines through those points which are further used for multiple decision-making process in later stage. 

In short, the main contributions made in this paper are as follows: (1) A novel concept of merging color based edge and depth edge to detect the boundary line on each side of the center point with which the robot gripper will make contact. That is the pair of line where the robot gripper touches the object for picking up. (2) The boundary lines serve multiple purposes: (2i) checking the angle between the boundary lines to conclude that the boundary lines form a valid handle (2ii) also we have formed a filter with the help of the boundary lines to detect
the obstacle on the way to pick that handle. (3) Finally, a cost function is being deployed to find out best possible handle (which minimizes the cost function) from a set of valid handles which already satisfied the criteria of (1), (2i) and (2ii). This handle will be the best handle based on the cost function. This handle will be used by the robot for grasping the object. Note that, the features for the cost function are based on the parameters of the handle (a) distance of the grasping region from the camera (b) angle between two boundary lines (c) the angle between the axis and those lines. This paragraph summarizes the main contribution of this work numbered as (1), (2) and (3). In the next few sections, we give detailed overview of each of (1), (2) and (3) and eventually validate the performance of the overall methodology. 
 
\section{Proposed Method} 
\label{sec:meth}
In this work, our main objective is to find suitable graspable affordances that can be grasped with the help of a two finger parallel gripper. The input to the pipeline is a color image and the depth point cloud obtained using a range sensor and at the end, the resulting grasp poses and position are fed to the robot to perform the grasping physically. Unlike other methods that captured input from multiple viewpoints and merged them to get a more precise view about the object to be grasped, we have extracted grasping regions from a single view frame only. In the literature, there are work experimenting with a various form of input such as- color, depth or the combination of both color and depth information. It is quite evident from their results that maximum performance can be achieved by combining both color and depth information rather than taking individually. We have also followed a similar procedure to develop a grasping algorithm that makes use of 2D image and 3D point cloud to reach derive a final graspable affordance. As our focus is to build a grasping algorithm that should be flexible enough to be applied in highly cluttered scenario found in a practical environment, we evaluated the proposed method in both single object and cluttered scene.

\begin{figure}
\centerline{
\includegraphics[width=8cm, height=3cm]{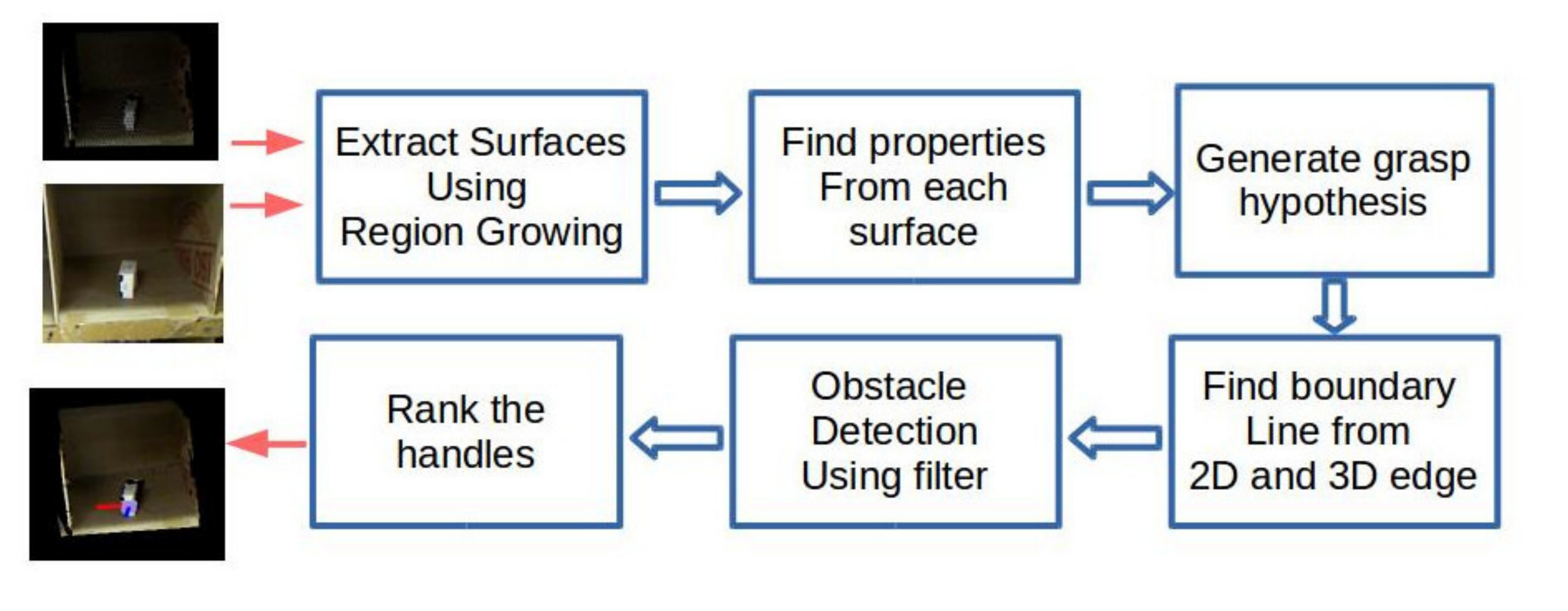}
}
\caption{ Flow Diagram of the proposed method}
\label{fig:flow_diagram}
\end{figure} 

A flow diagram of the proposed method is shown in Figure \ref{fig:flow_diagram}. The pipeline first started with the creation of multiple surfaces in the point cloud based on a smoothness criterion extracted using region growing algorithm\cite{rabbani2006segmentation} in 3D. This enables us to distinguish objects from each other using depth information.
This algorithm \cite{rabbani2006segmentation} checks the angle between a seed point and its neighbors. If the angle is less than a threshold then it adds the point in the current region and starts growing from the point. We group points that have similar characteristics in 3D to form a segment. At the end of this process by grouping points together we end up getting a few different segments. Then for each segment, we calculate some features which will be helpful for finding the grasping location in the next level of the pipeline. Now comes the main part of our pipeline where we have applied a two-stage process to find the grasping handles. To produce a high performance the precision of the grasping algorithm must be very high, otherwise, the object may fall during picking. The recall does not have much effect on the performance as even with low recall there will be plenty of handles to choose from. So, here we have focused on achieving high precision. We report the quality of the performance in terms of precision.
After grouping points with similar characteristics into a surface, we find out the two dominant directions associated with each surface by Principal Component Analyses (PCA). This helps us to find the orientation of the surface in 3D as a warm start so that will have some direction to go forward with. These two directions and the normal of the surface are important for our algorithm as the physical gripper of the robot hand will align along these directions. For finding handles for a successful grasp, some basic criteria are needed to be fulfilled. A valid handle will be the one which has sufficient gap around it and also the robot hand will not collide with any other object while picking using that handle. Also, the object should be stable inside the gripper and will not fall during the process. Towards this goal we first select some sample grasping locations based on the first criteria i.e having a necessary gap around the potential handle. Among those hypotheses, we again check the stability of the handle by extracting two parallel lines with whom the gripper plate will make contact. If the two lines are not parallel to each other then the object cannot be picked up by the parallel gripper plates. Hence, rejecting the hypothesis. In the last stage, we want to make sure that the robot hand will not collide with the gripper while approaching the handle. For this, a filter is formed using the two parallel lines and is being checked for any potential occlusion. The remaining hypotheses that satisfy the above criterion are included in the set of valid handle and grasping can be done using any one of these handles. However, for physical grasping, only a single handle is required at a time. Instead of randomly choosing a handle from the set, a cost function is formulated based on the features we get from the handles. The handle for which minimum cost is obtained is used for grasping.

In summary, we have started with color and depth information of workspace from a single viewpoint. These features then go through multiple steps that finally lead to grasping pose detection. The whole pipeline does three main task,- (1) creating multiple surface segment inside the bounding box from object recognition module, (2) identify grasp handles with poses using those surfaces and finally (3) rank the handles to choose the best one. Each step of the pipeline is described in detail in the next section.

\subsection{Segment the cloud into patches}
The input of this step is an image and a point cloud of the workspace obtained from a single view. The result from this step is we obtain some The input of this step is an image and a point cloud of the workspace obtained from a single view. The result from this step is we obtain some surfaces that belong to the object. Each continuous surface is a group of points that share similar characteristics. We can compare the problem of finding a grasping location with object detection, where the object to be detected can be of any size and anywhere in the frame. In general, in the past, this kind by problems was approached by searching for objects in multiple scales with the help of sliding windows. For object detection, the processing is done in the 2D image, whereas, grasping is a 6DOF problem. If we want to apply the technique of sliding window for the 6DOF problem, the computational and time complexity increases in a huge amount making the method intractable for any real-time processing. This problem is addressed in the literature by considering depth information like RGB information making depth data as another channel in 2D (give references of deep learning). Some work \cite{Pas2013LocalizingGA}\cite{Pas2015UsingGT} randomly sample a few points from the workspace at a fixed scale and check whether the robot hand can pick the object using that handle. So, in order to make the searching process easier and effective we first group similar type of points into patches and extract properties from each patch that in turn helps to find the grasping handles. 

\textbf{Region Growing-}
This grouping is done by region growing algorithm \cite{rabbani2006segmentation} where multiple segments are extracted by grouping the points which belong to the same smooth surface. The decision of grouping the points are taken based on a threshold value. The quality of the surfaces is hugely dependent on the threshold value.  But as the input to our pipeline is noisy 3D input data, region growing leads to over-segmentation if the threshold is set too high. So, the threshold is set in lower range and by our experiments, the surface are reasonably good. From each surface a set of parameters i.e centroid, average surface normal and first two dominant directions obtained from Principle Component Analysis (PCA) are extracted which will be used in later stage.
  
%

\section{Grasping}
\label{sec:grasping}
In robotics automation, grasping is an important task. It enables robots to manipulate the environment as done by the human beings. A lot of research has already been done in this domain in the quest to make human-like grasping modules, even though the desired performance has not been achieved yet. Through this work, we have devised a novel grasping methodology which yields very high success rate for a successful grasp. First of all, for a successful grasp of an object, it is required to know the physical structure of the robot hand. In this work, we deal with the simplest form of robot gripper which is a two finger parallel gripper. Another major requirement for a successful grip is that the distance between the two grippers is needed to be sufficient enough so that the gripper could be placed around the handle to pick the object successfully. Let the thickness of the gripper plate be denoted by t and the minimum gap on either side of the handle be denoted by $g$. For a successful grip we need to have the following criteria to be satisfied:

\begin{enumerate}
\item For a successful grip, we need to have $g > t$.
\item The robot hand should not collide with other objects while approaching the specified object
\item The boundary at either side of the handle should be as parallel as possible to each other.
\end{enumerate}

Criterion (3) ensures that if the handle boundaries are parallel then the portion of the handle that makes contact with the gripper is more stable and hence reduces the chance of the object falling down. These boundary lines also help in creating a mask which helps us to validate criterion (2). The algorithm proposed in this work, first validates criterion (1) and then only it proceeds to validate the other two criteria.

\begin{figure*}[!t]
  \centering
  \begin{tabular}{cc}
\includegraphics[width=0.3\linewidth,height=0.25\linewidth]{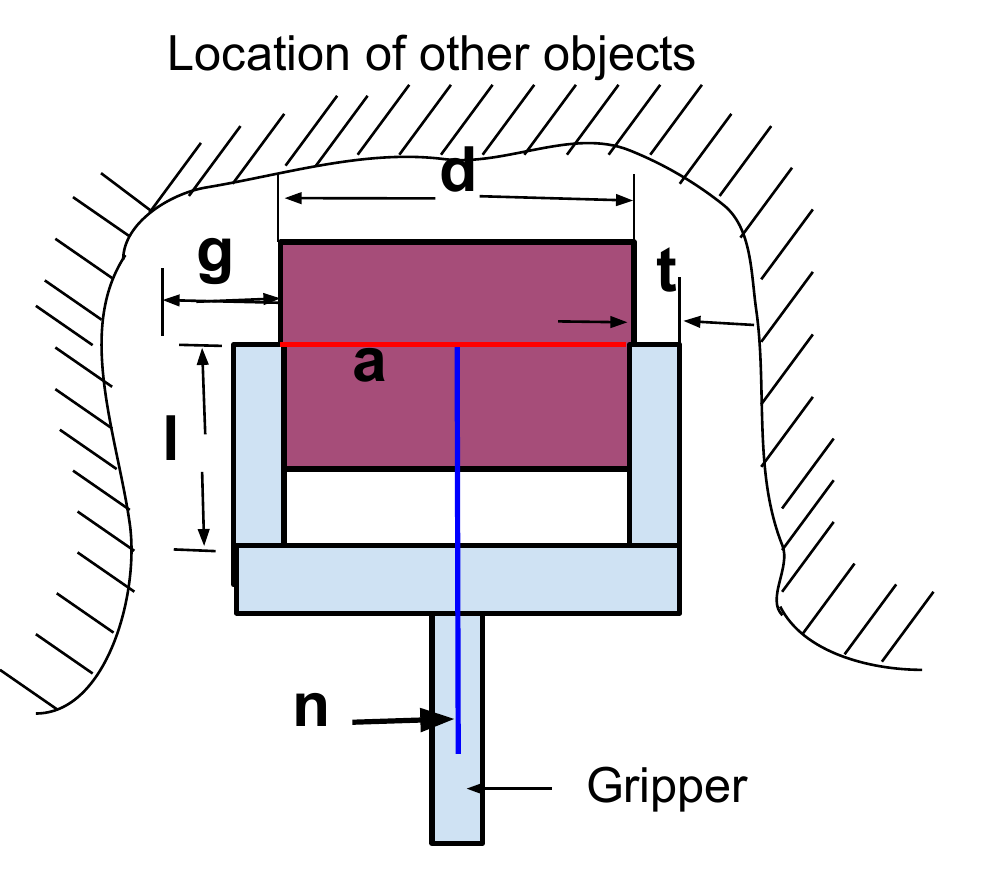} &
\includegraphics[width=0.3\linewidth,height=0.25\linewidth]{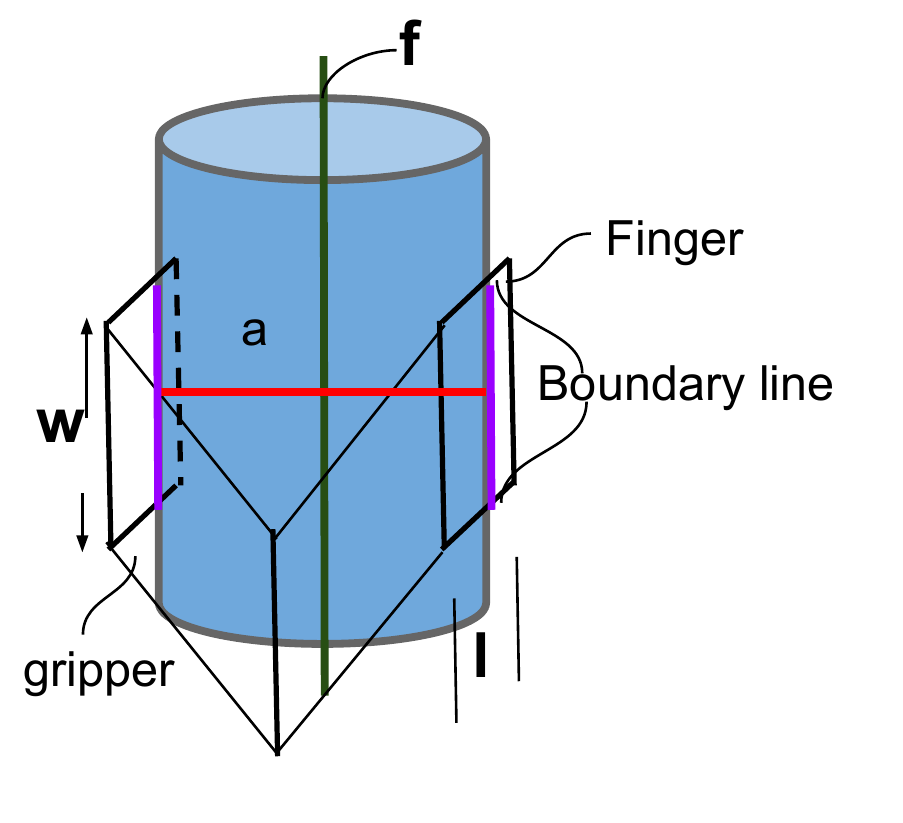} \\
    \scriptsize{(a) Gripper Geometry and Clearance between objects} & \scriptsize{(b) Boundary lines} 
  \end{tabular}
  \caption{Grasp Configuration for a two finger parallel jaw gripper.
    For a successful grasp, the clearance between objects must be
    greater than the width of each finger ($g > t$). The gripper
    approaches the object along a direction opposite to the surface
    normal ($\hat{n}$) with its gripper closing plane coplanar with the axis $\hat{a}$
  of surface segment as shown in (a). Two boundary lines in either side of the center are shown in (b) with which the gripper plate will make contact during physical grasping. For stable grasp these two line should be parallel to each other.}
  \label{fig:gripper_object}
\end{figure*}

To give a detailed overview of the algorithm we first define a few notations which will be used throughout this write-up. The proposed method takes 3D point cloud $\mathscr{C} \in \mathscr{R}^3$ and physical geometry of the gripper and generates a 7DOF grasping pose and location $h$. $\mathscr{H}$ is the set of all valid handles. Some of the notations are borrowed from the paper\cite{Pas2015UsingGT}. The geometry of robot hand is defined by $\theta = (l,t,w,d)$ where $l$ is the length of gripper fingers, $d$ is the maximum opening by the gripper fingers, $w$ is the finger width and $t$ be the thickness of the gripper. As stated earlier $g$ has be greater than $t$. The gripper closing plane ($\hat{a}$) is defined by the plane of the finger motion along which we have to find the gap $g$. Along with $\hat{a}$, the robot approaching direction $\hat{n}$ is used to define a the direction of the handle. Now if the center of the handle is given by $c = (x,y,z)$ where $ c \in \mathscr{R}^3$, then a handle $h$ is represented by $h = (c,\hat{f},\hat{n},\hat{a},r)$ where $\hat{f}$ is normal to both $\hat{a}$ and $\hat{n}$ and $r$ is the radius of the handle.

\subsection{Finding The Gap}
As discussed in Section \ref{sec:grasping}, we will first search for the required gap (g) around a point. For this, we will be using the same methodology as described in the paper \cite{Pas2015UsingGT} where they took a fixed radius sphere around a point and formed a Darboux frame with the help of surface normal and principal curvature directions as calculated by a Taubin quadratic surface fitting method. But, in \cite{Pas2015UsingGT} they do not explore surface discontinuities and hence it decreases the overall performance accuracy. In our work, we enhance the accuracy of the performance by exploiting the discontinuities on the surface. Using the discontinuities of the surface we get an approximate idea of the spread of the surface which produces more accurate principal directions with a higher precision. So, in our work, the three axes of Darboux frame are formed by the surface normal, and one major and a minor axis which is derived using Principal Component Analysis (PCA). Once, the Darboux frame is formed, we check whether along y-axis the handle is collision free for some distance. If this condition is satisfied, then the handle is added to the hypothesis set (H). This nominal enhancement already outperforms the existing methodology in terms of accuracy but the accuracy rate is still not high enough. The reason, for that, is, the region growing algorithm does not produce perfect segmentations due to noisy real-life input data. Due to the presence of noise in the data, the major and minor axis obtained through PCA are also not precise enough. Even though, the faulty major and minor axis computed from the noisy data may still produce a gap (g) which satisfies criterion (1), the overall precision for a successful grasp decreases. 

\begin{figure*}
  \centering
  \begin{tabular}{cc}
   \scalebox{0.2}{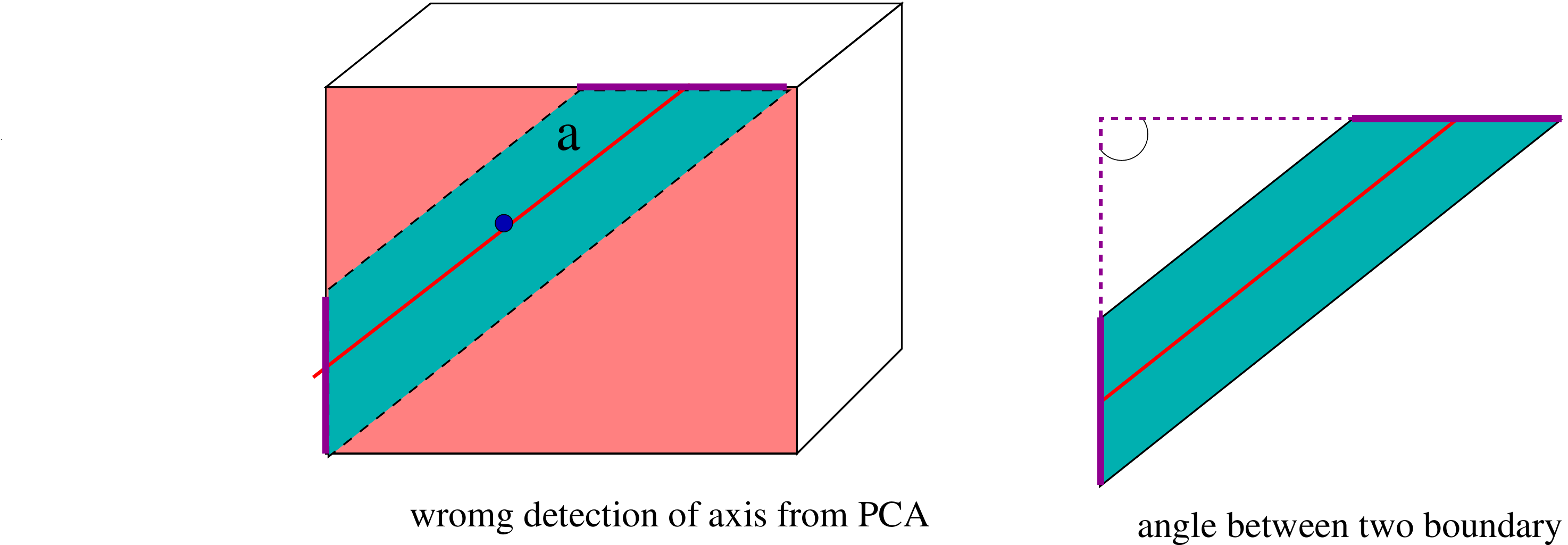} &
   \scalebox{0.15}{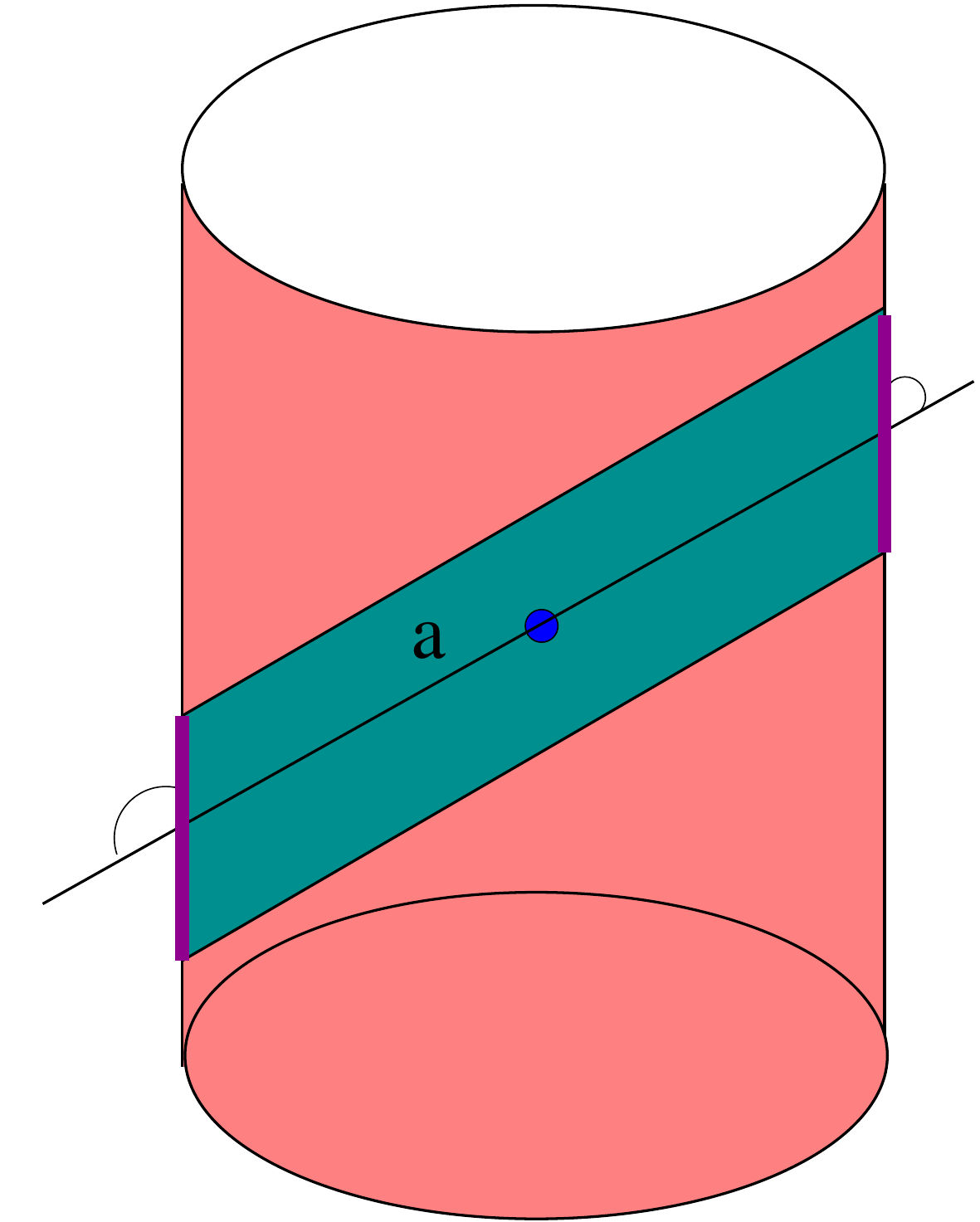} \\
    \scriptsize{ Two example of erroneous axis calculation in (a) box } & \scriptsize{(b) cylinder} 
  \end{tabular}
  \caption{ Two example of erroneous axis calculation shown in box (a) and cylinder (b). In case of box the particular direction of axis $\hat{a}$ give us two boundary line ($l_{+}$ and $l_{-}$)that has a angle $\theta_b$ which cannot not be picked by the parallel gripper. For cylinder, as shown in (b), even if the boundary is parallel to each other for the axis direction, but still that is not a valid grasping pose as the boundary lines and axis $\hat{a}$ are not normal to each other ($\theta_{axis} \ncong 90^\circ$).}
  \label{fig:error_pca}
\end{figure*}

Two example of such behavior are given in Figure \ref{fig:error_pca}.  In the first figure (a), the front surface of a box is tested to detect a valid handle. The major axis from PCA in this case is not along the box length. With such direction of $\hat{a}$, the earlier stage detects the required gap $g$ in both side leading to a false detection. Also, another example is shown in figure (b) where the $g$ is present but the pose information is incorrect.

\subsection{Finding two boundary lines} 
If we encounter some condition as the first case of Figure \ref{fig:error_pca} i.e picking the box using the handle, then the first stage will include the region in grasp hypothesis set $\mathscr{H}$. But when we try to pick that box using the handle, it will slip from the gripper. This is because we have used parallel finger gripper and, to form a tight grasp the two edges of the handle along the axis $\hat{a}$ has to be parallel to each other. In this way the gripper can get a good hold of the object and perform a stable grasping operation. So, in this section, we have tried to find out two contact lines at either side of the center point defined by $l_{+}$ and $l_{-}$ where $l_{+}$ be the boundary line along the axis $\hat{a}$ and $l_{-}$ is the boundary line located on the other side. The angle between these two lines decides the validity of the handle. 

The ideal way to find out two line with whom the gripper plate would made contact is to extract the edges on each side and fit two straight lines through the points. With synthetic 3D data it can be done easily by extracting the edge points from the already found handle and fit a line through it. But as here we are dealing with noisy point cloud data and the noise often removes some points from point cloud which leads to detect false edge or remove valid edges. Another kind of edge exists in 3D\cite{depthEdge} whom we can take into consideration as a method of valid edge detection. But again, due to spurious nature of the point cloud data it may lead to confusion. If we do not want to deal with noisy 3D sensor input we can rely on 2D color based canny edge detection edges. But image based edge detection depends on illumination and there may be edges inside the handle due to color changes. And also if the color of the object is same as background then there will be no edge corresponding to the boundary. Ideally 3D edge points is more reliable than 2D edge points in case of detecting boundary edge points. It is evident from our experiment that if color or depth information is used separately, the expected performance would decrease rapidly. Like other existing work, we also decided to merge this two information to get reliable boundary edge points that would be used to fit lines.  

A 2D or 3D edge may be spurious due to color variation or noisy depth data, but if these two type of edges appear in the vicinity of each other we can conform that one is a reliable edge. Again out of all reliable edges. we have to filter out the ones corresponding to two boundary.
Towards that goal, we take help of handle edge points and if, for a handle edge either canny or depth edge exists, then we add the points to boundary edge set. In the next paragraph, this process is described and at the end of it we get two set of points to fit st. lines.

\begin{figure}[!t]
\centering  
\subfigure[]{\label{fig:a}\includegraphics[width=0.2\linewidth,height=0.15\linewidth]{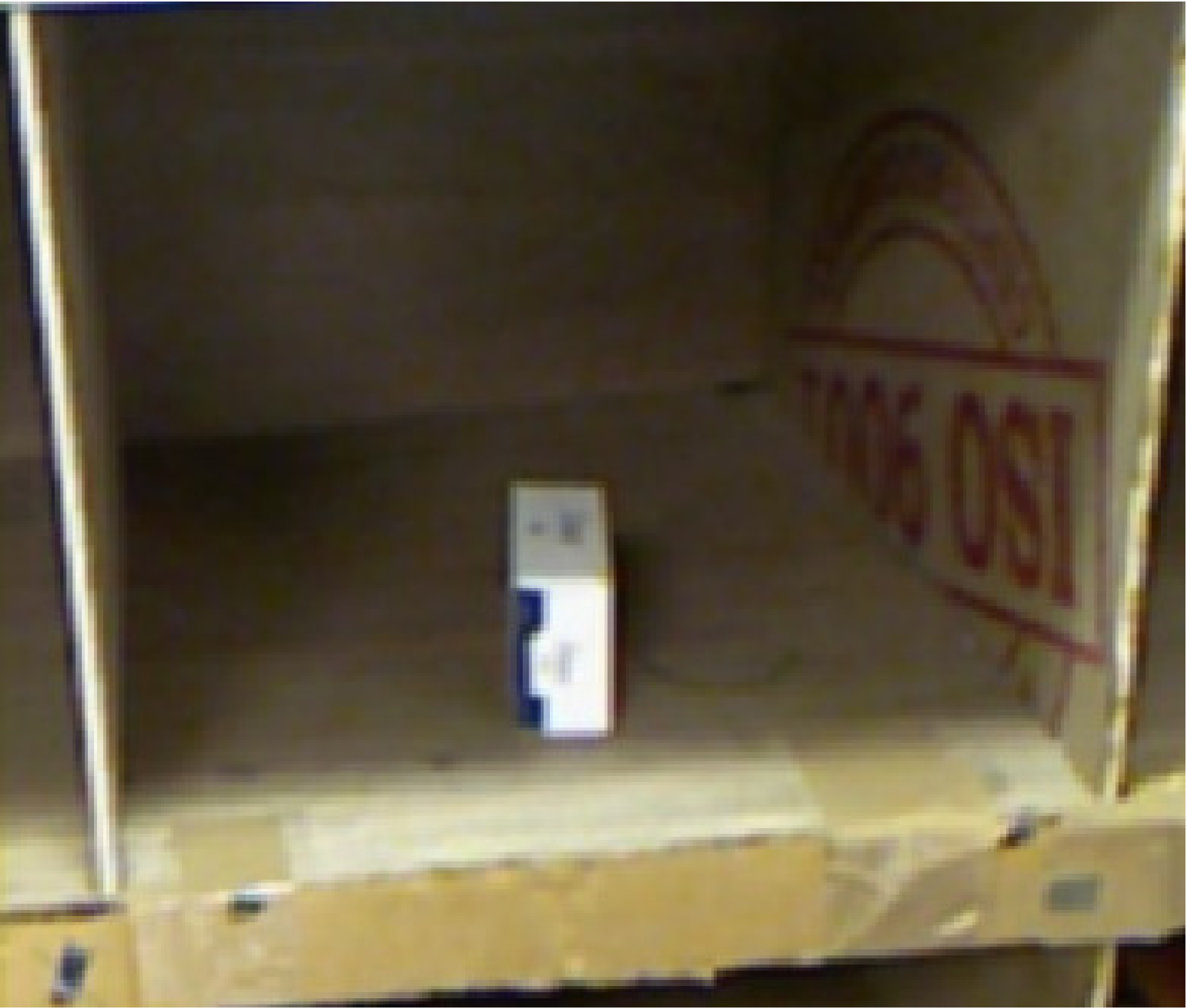}}
\subfigure[]{\label{fig:b}\includegraphics[width=0.2\linewidth,height=0.15\linewidth]{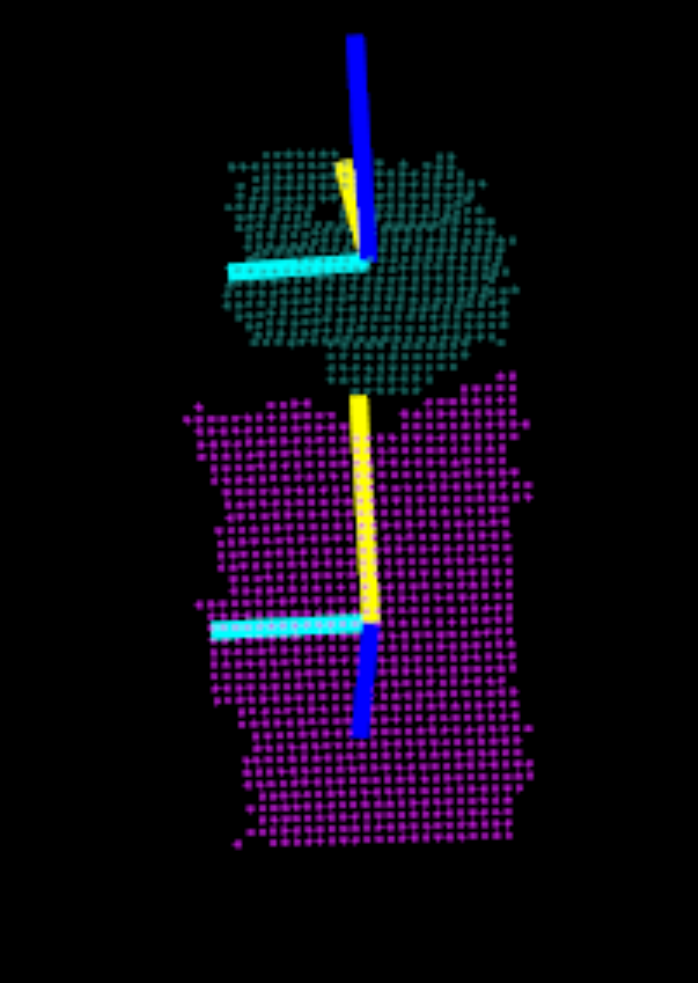}}
\subfigure[]{\label{fig:c}\includegraphics[width=0.2\linewidth,height=0.15\linewidth]{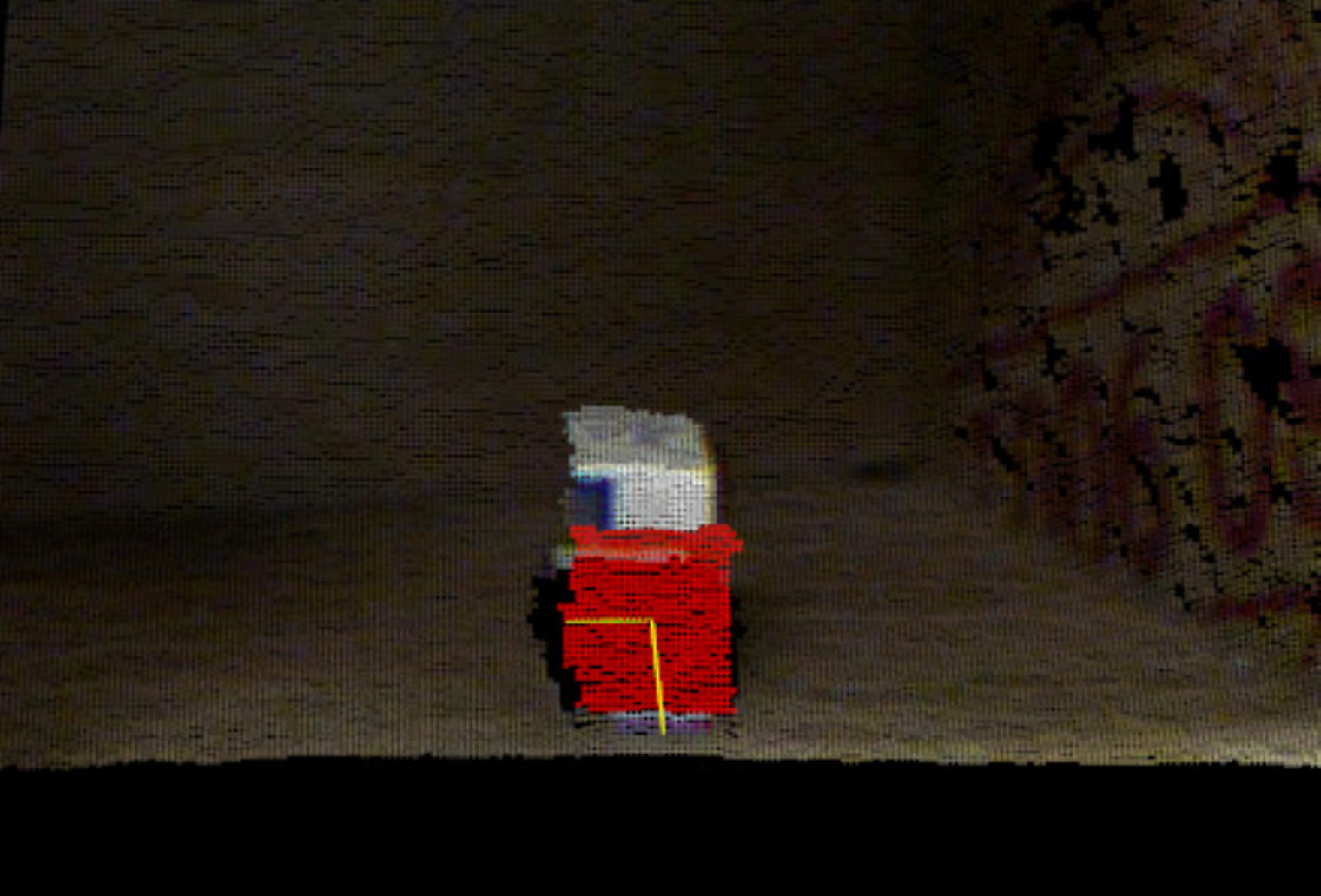}}\\
\subfigure[]{\label{fig:d}\includegraphics[width=0.2\linewidth,height=0.15\linewidth]{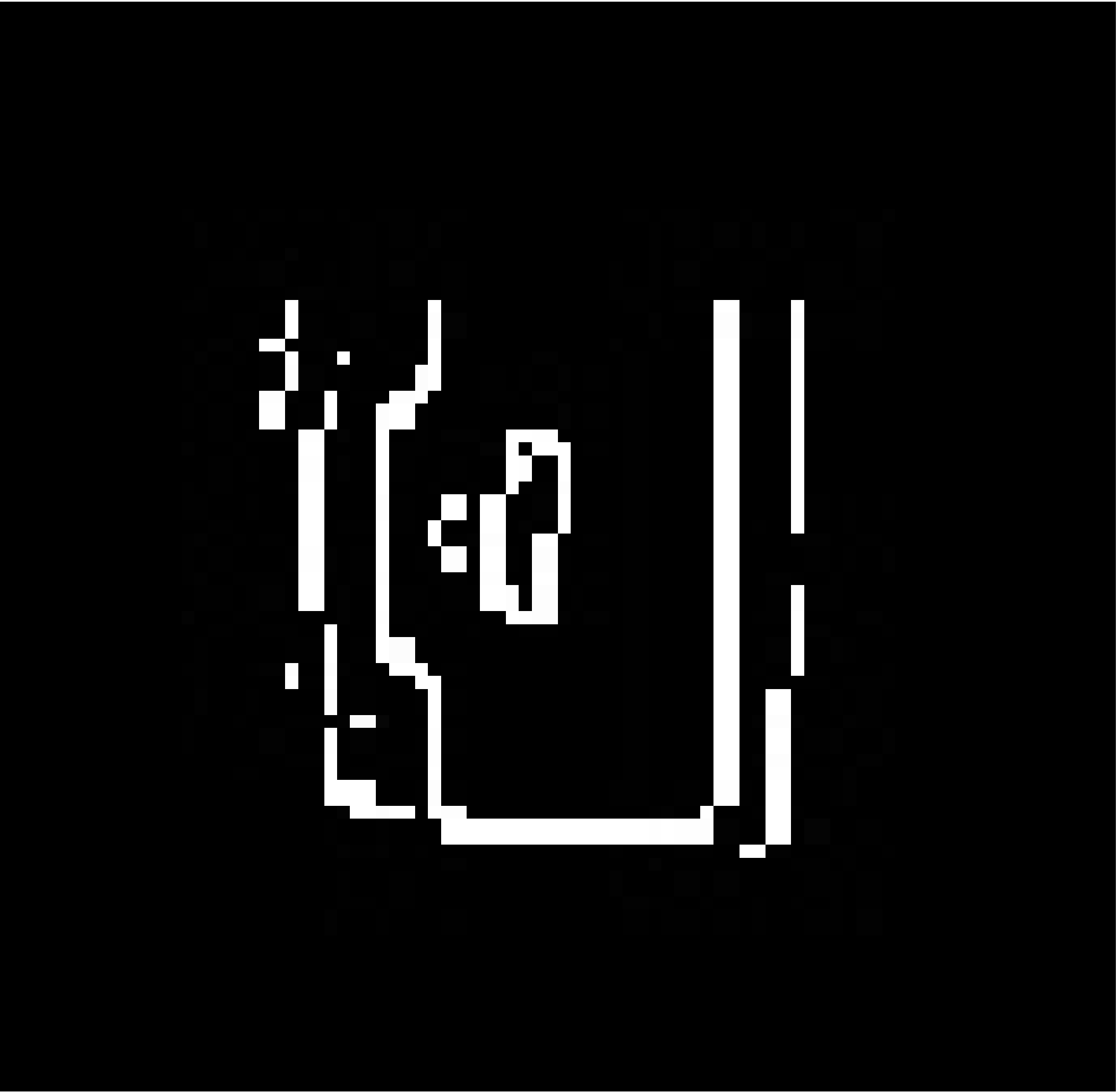}}
\subfigure[]{\label{fig:e}\includegraphics[width=0.2\linewidth,height=0.15\linewidth]{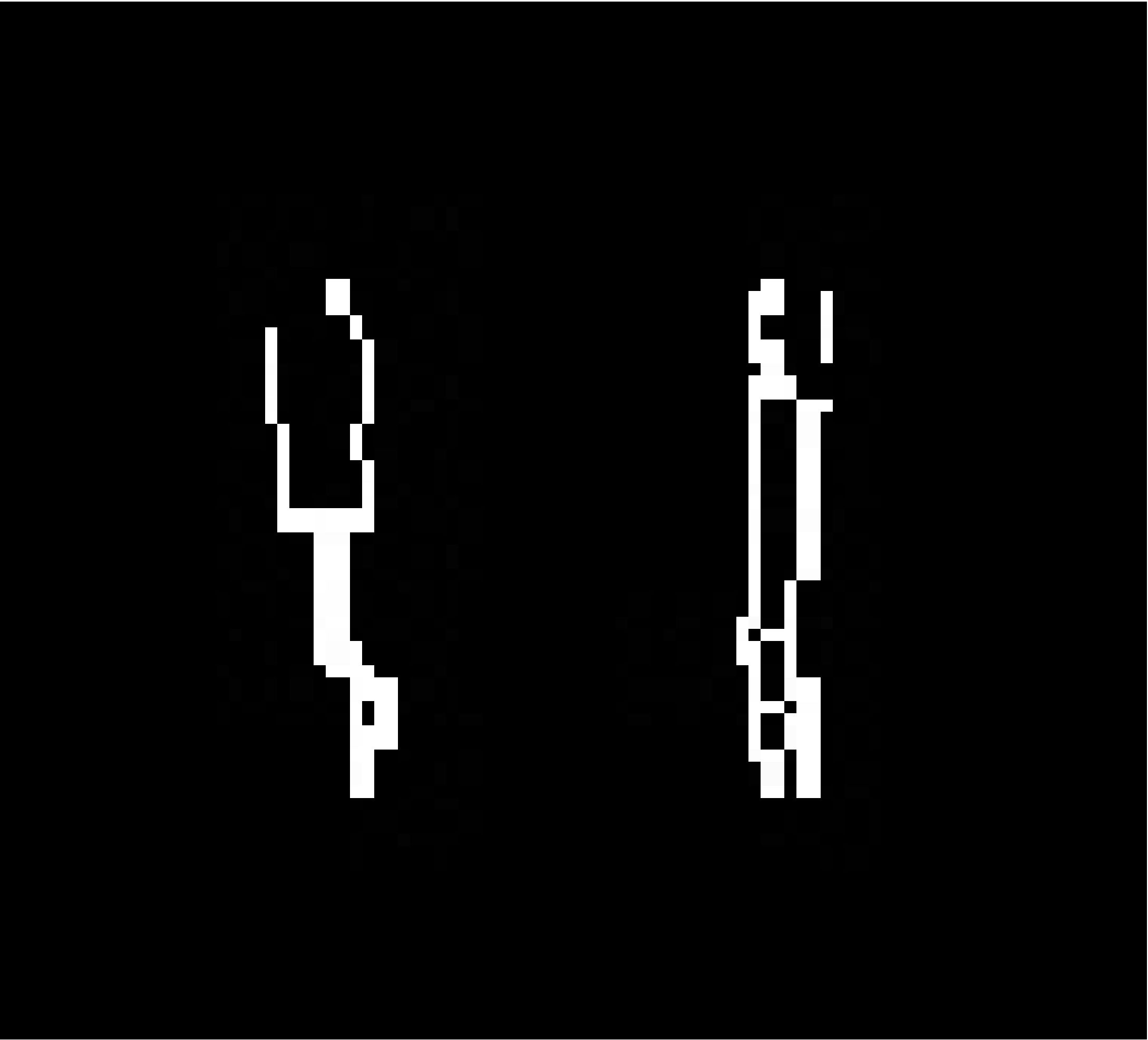}}
\subfigure[]{\label{fig:f}\includegraphics[width=0.2\linewidth,height=0.15\linewidth]{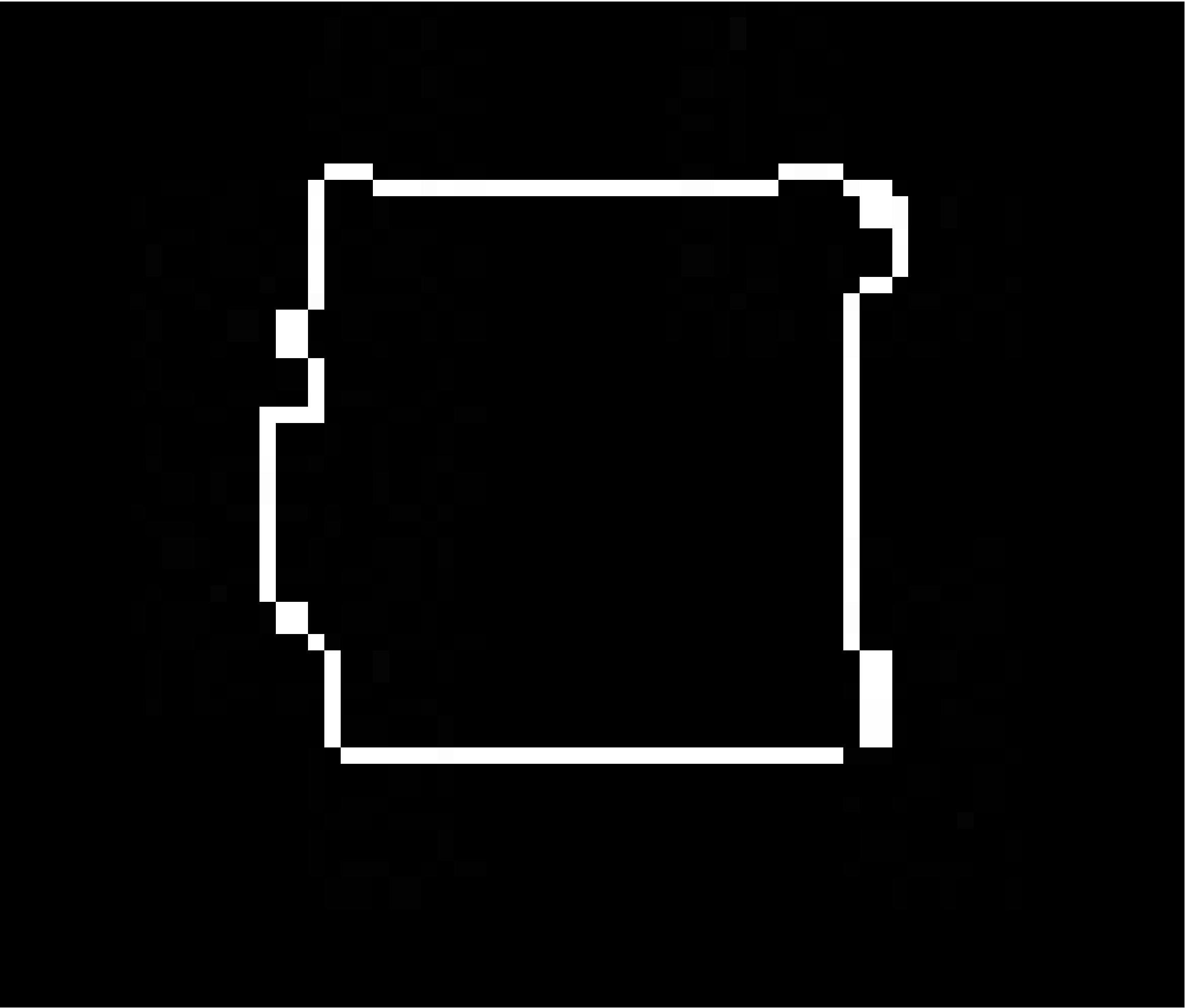}}
\subfigure[]{\label{fig:g}\includegraphics[width=0.2\linewidth,height=0.15\linewidth]{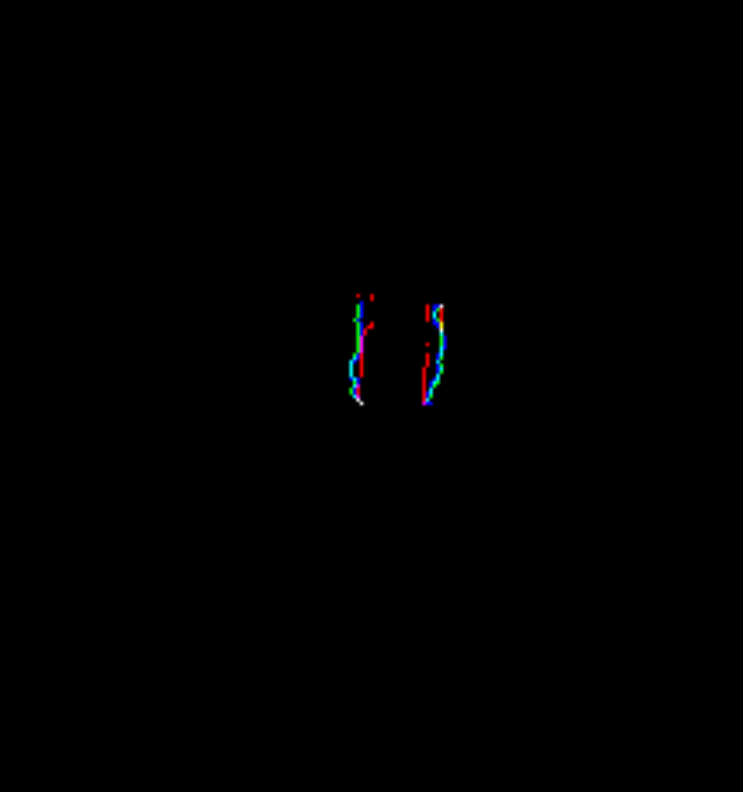}} \\
\subfigure[]{\label{fig:h}\includegraphics[width=0.2\linewidth,height=0.15\linewidth]{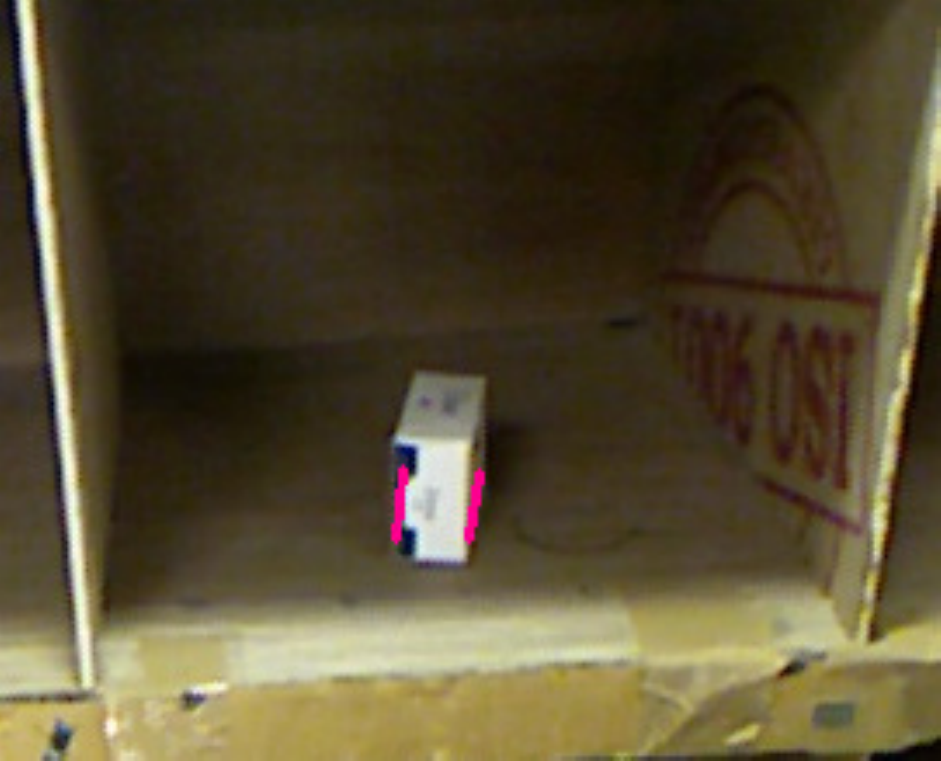}}
\subfigure[]{\label{fig:i}\includegraphics[width=0.2\linewidth,height=0.15\linewidth]{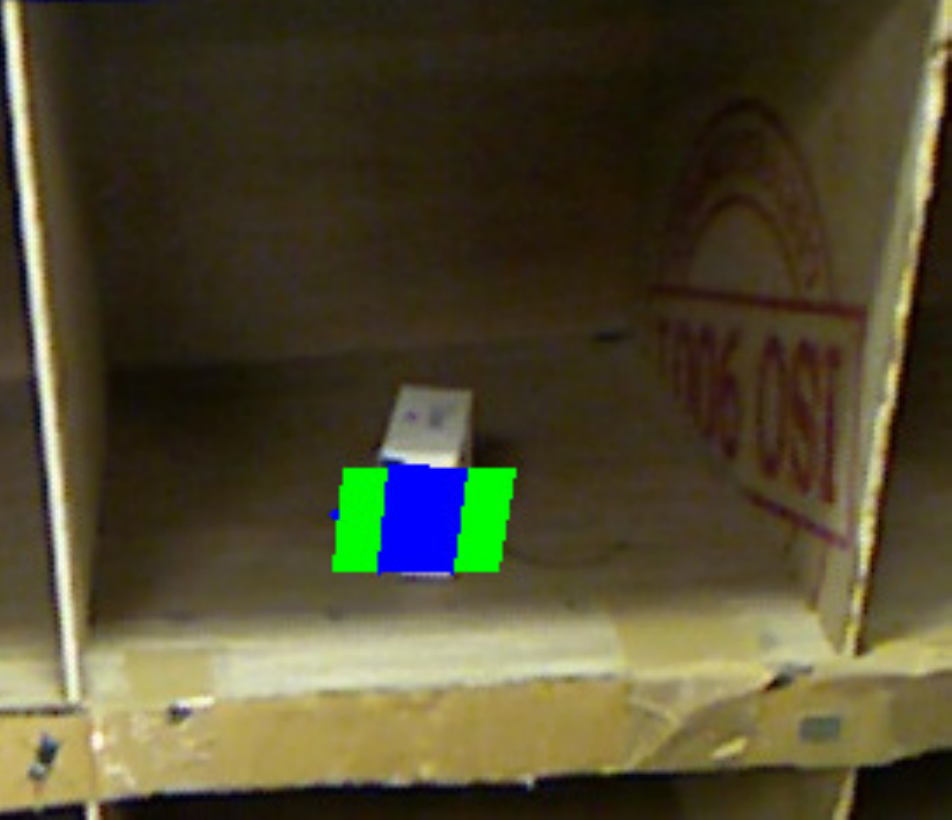}}
\subfigure[]{\label{fig:j}\includegraphics[width=0.2\linewidth,height=0.15\linewidth]{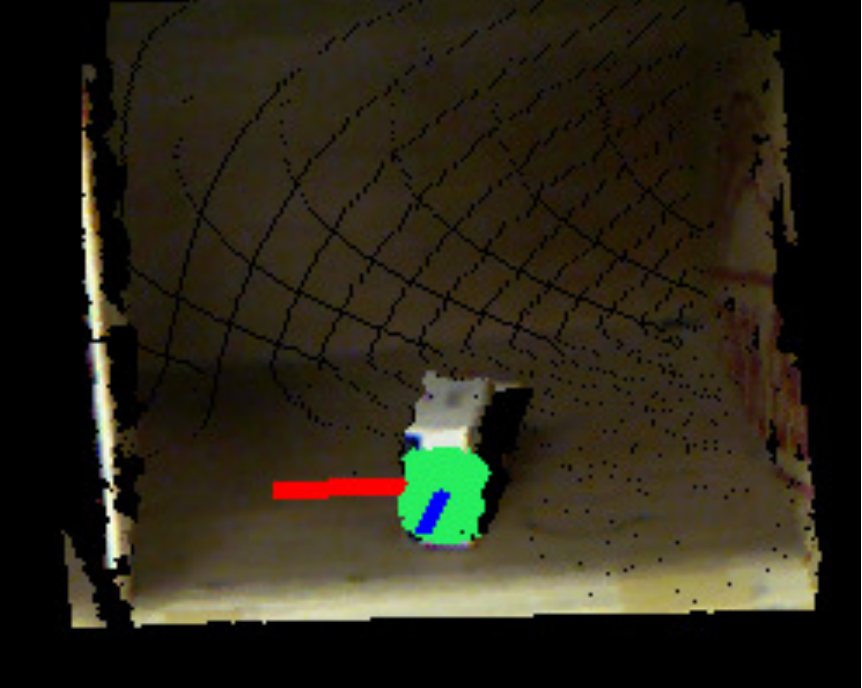}}\\
\caption{ First two are input image and region growing surfaces with PCA major and minor axis. Third one is grasping region found by our previous method and last figure is canny edge detected image of a particular bin. The bounding box shown in Violet color in the canny image is calculated from the grasping region found (third figure).}
\label{fig:bounadry_line}
\end{figure}

At first, boundary points corresponding to the handle is extracted and then those points are validated using 3D depth edge and 2D canny edge points.
Let $E_c$ be the set of edge points we get from canny edge detection in color image and $E_d$ is the 3D depth edge points based on surface normal as defined in the work[]. For each hypothesis $h \in \mathscr{H}$ the corresponding boundary points are defined by $E_s$ (as shown in figure\ref{fig:bounadry_line}). Let $E_{v}$ be the set of all reliable boundary points.
Now, if a 2D ($e_c$) edge is present corresponding to an boundary point $e_s$ then we keep that edge in valid boundary set $E_v$. Sometimes, a 2D edge may not exist due to same color but in depth space there may be a discontinuity. So for a point if a depth edge exists then also that point is retained as boundary. All the points for which neither $e_c$ nor $e_d$ exist, are removed. After this process, we get two set of points corresponding to each boundary and apply a line fitting algorithm that returns the center point and slope of the line. Let $L_i$ be the level of each point in set $E_s$, then- 

\begin{equation}
   L_i =
    \begin{cases}
      1, & \text{if}\ e_{c_i} \in \left \{E_c\right \} \ \ or \ \ e_{d_i} \in \left \{E_d\right \}\\
      0, & \text{otherwise}
    \end{cases}
    where \ i = 1,\dots, size(\left \{E_s\right \})
    \label{eg:level}
\end{equation}

\begin{equation}
\left \{ E_v \right \} \leftarrow e_{d_i}\ or\ e_{c_i} | \ L_i = 1 \ \ i = 1,\dots,size(\left \{E_s\right \}) \}
 \label{eq:rs}    
\end{equation}

This can be explained further from the figure \ref{fig:bounadry_line}. The figure \ref{fig:a} presents the object to be grasped. Two surfaces from region growing algorithm corresponding to the object is shown in \ref{fig:b} along with major, minor and normal direction. The front surface is tested for the first criteria and it detects a sufficient amount of gap around the surface along major axis direction (\ref{fig:c}). After it gives a positive result, we go ahead with the second stage. The figures in the second row represents canny edge, depth edge and surface edge respectively. \ref{fig:g} is the merged image containing three type of edge. This filtering and merging has been done according to equation \ref{eg:level} and \ref{eq:rs}.  We can clearly differentiate two set of boundary points on each side of center point.
Finally, two fitted line through the points at the boundary position is shown in the figure \ref{fig:h}. These are the lines with which the gripper plates are supposed to make contact while physical grasping. \ref{fig:i} The last figure shows the mask that detect any potential occlusion with other objects.

The line fitting equation returns a direction vector $u = (u_x,u_y)$ and point $(x_c,y_c)$ on the line and the equation of the line is,-
\begin{equation}
l_{\pm}(x,y) = (x_c,y_c) + t*(u_x,u_y)    
\end{equation}

At first, the angle between both the lines is checked. This is done by checking the dot product between the lines $l_{+} \&\ l_{-}$ Ideally the value should be zero, but for practical purpose if the angle is greater than a user defined threshold then that handle is rejected. 

\begin{equation}
  if\ (u_{+}\cdot\ u_{-}) > \theta_r,\ then \left \{\mathscr{H}\right \} \leftarrow \left \{\mathscr{H}\right \} / h
\end{equation} 
After this, a last stage of rechecking is done using the boundary lines so that the robot hand does not collide with any other object on the way to pick using that handle. This is done by forming a filter as shown in figure \ref{fig:i}. Those handles for which the filter detect obstacle in the front, are rejected. This step is required, as at first we restrict our workspace within a sphere around the handle. This filter takes care of points that is outside the sphere.

\subsection{Rank the handles}
Now, we have the set of valid handles any of which can be used to pick the object by the robot hand. But only one handle is required to carry out the grasping process. Rather than choosing one randomly from the set of handles, we formulate a cost function $f_c$ that will assign
some score to each handle based on their properties. These properties are chosen from the handles we have already found out. As per earlier discussion, ideally the pair of boundary line ($l_{\pm}$) should be parallel to each other. But as we are dealing with real life noisy data, all the boundary lines with little deviation is considered as a valid handle. For ranking, this angle is taken as one of the decision making factor where lesser the angle the better the ranking is. Also, the axis of the handle has to be perpendicular to the gripper finger, therefore the angle between the axis and boundary lines is considered as another property of the cost function. Ideally the angle should be around $90^\circ$.
This two angle are represented by corresponding dot product in cost function ($f_c$). Hence any value close to 0 gets a higher ranking because the dot product between the axis and the average direction of the two boundary lines having an angle of 90 degrees i n between them has a dot product to 0. Again, the handle which is located close to the camera should be given higher priority compared to the one that is situated far away. Using these three features, we did the ranking of the handles and for evaluation purpose top five handle is returned.

 $f_c$ is the cost function to be minimized. $a_b$ is the dot product between the two boundary lines and $a_{axis}$ be the dot product of axis and average of boundary lines. To be a perfect handle, the value of $a_{axis}$ should be close to $0$ and $a_b$ to $1$. Both the range of $a_b$ and $a_{axis}$ is $\left \{0,1\right \}$. As we chose to minimize the cost function $f_c$, the value of $a_b$ is flipped by subtracting it from $1$. Let us define $c_z$ as the distance of the center of the handle. To use $c_z$ in the cost function, it is normalized to $\left \{0,1\right \}$, $0$ being min distance handle from the input device. The cost function is defined by,-

\begin{equation}
\begin{split}
f_c^h  ={} & \ w_1 \ a_b +\ w_2 \ a_{axis}+\ w_3 \ c_z \\
& where, h = 1,\dots,size(\left \{\mathscr{H}\right \}) \\
& w_1,w_2,w_3 \ =  the\ weights\ associated\ with\ each\ feature \\
& \ a_b\ = 1 - (u_{+}\cdot\ u_{-}) \\
&\ a_{axis}\ = (a\cdot\ u_{\pm})\\
&\ c_z \ = \ (c_z - w_{zmin})/(w_{zmax} - w_{zmin}) 
\end{split}  
\label{cost_func}
\end{equation}

Ranking of the handle is done by sorting the cost function in ascending order. Less the $f_c$ better the ranking. So, the handle for which $f_c$ gives minimum value gets first chance to be picked by the robot. If it fails, they we proceed with the next one.

\section{Simulation and Experimental Results} 
\label{sec:res}
In this section, we provide results of various experiments performed to establish the usefulness of the proposed algorithm in comparison to the existing state-of-the-art methods.  The input to our algorithm is a 3D point cloud obtained from Kinect \cite{zhang2012microsoft}/ realsense \cite{draelos2015intel}/Ensenso \cite{ensenso} and, the output is a set of graspable affordances and pose. An additional smoothing pre-processing step is applied to the point cloud obtained from Ensenso which are otherwise quite noisy compared to that obtained using either Kinect or realsense sensors. The performance of the proposed algorithm is compared with other methods on four different datasets, namely, (1) Cornell Grasping dataset \cite{lenz2015deepgrasp}, (2) ECCV dataset \cite{Aldoma2012}, (3) Kinect Dataset \cite{SegIROS11} (4) Willow garage dataset \cite{willow}. As most of the dataset contain enough cluttered scene, we have tested our method on some household objects in both single object and cluttered scenario.

Different authors use different parameters to evaluate the performance of their algorithm. For instance, authors in \cite{plattgrasppose2016} use \emph{recall at high precision} as a measure while few others as in \cite{lenz2015deepgrasp} use \emph{accuracy} as a measure. The precision is usually defined as the fraction of total number of handles detected which are true.  There are other researchers as in \cite{pas2015using} \cite{jain2016grasp} \cite{vezzani2017grasping} who use \emph{success
rate} as a performance measure which is defined as the number of times a robot is able to successfully pick an object in a physical experiment. The success rate is usually directly linked to the precision. As we want our method to run with real life manipulation system, precision will give the most relevant performance measure for such system. Because, even with low recall there will be a plenty number of option to choose from, but low precision may lead failure of the grasping process therefore the performance will degrade drastically.

We demonstrate the performance of the proposed algorithm in picking objects from both cluttered and single object scenario. The grasping algorithm comprise of multiple stages to get as perfect performance as possible. So, we carefully observed the effect of each step on the final precision. Table \ref{tab:indobj} shows the improvement in performance for single objects after the completion of each stage of the pipeline. Only the first stage i.e generation of grasp hypothesis itself results around 85\% precision in our experiments. Addition of the criteria of rejecting all hypothesis for which the boundary line is not parallel to each other improves the performance upto 92\%. And the overall precision of the system is 94\% after the addition of one final obstacle detection stage. The hypothesis generation stage of paper \cite{Pas2015UsingGT} gives 73\% success rate, where ours generates 85\%. This increase of precision happens due to the fact that they randomly choose a point with a fixed radius as a potential grasping region, whereas we first explore surface discontinuity to mark the end of a region. This simple step boosts our performance by a huge amount and make the system flexible enough that it can accommodate handles with different sized radius. The overall precision is also improves significantly from $88\%$ to $94\%$.

\begin{table}[t]
\centering
\caption{Precision on Individual Objects}
\label{tab:indobj}
\scriptsize
\begin{tabular}{|c|>{\centering\arraybackslash}m{1.3cm}|>{\centering\arraybackslash}m{1.5cm}|>{\centering\arraybackslash}m{1.3cm}|}
  \hline
   &    \multicolumn{3}{c|}{\% Precision} \\ \hline
  Object     & First Stage &After parallel line &Overall \\ \hline
Toothpaste    & 86      & 94         & 96         \\ \hline
Cup           & 85      & 90         & 91         \\ \hline
Dove Soap     & 88      & 94         & 96         \\ \hline
Fevicol       & 86      & 94         & 95        \\ \hline
Battery       & 88      & 93         & 95        \\ \hline
Clips         & 75      & 83         & 86          \\ \hline
CleaningBrush & 94      & 96         & 100         \\ \hline
SproutBrush   & 80      & 90         & 92         \\ \hline
Devi Coffee   & 82      & 92         & 95         \\ \hline
Tissue Paper  & 86      & 93         & 95         \\ \hline
Total         & 85     & 92           & \textbf{94}     \\ \hline
\end{tabular}
\end{table}

\begin{table}
\centering
\caption{Performance Comparison for Individual Objects (Precision)}
\label{tab:indobjcomp}
\scriptsize
\begin{tabular}{|c|>{\centering\arraybackslash}m{1.3cm}|>{\centering\arraybackslash}m{1.5cm}|>{\centering\arraybackslash}m{1.3cm}|>{\centering\arraybackslash}m{1.3cm}|}
  \hline
   &    \multicolumn{4}{c|}{\% Precision} \\ \hline
   &    \multicolumn{2}{c|}{Grasp Hypothesis Generation} & \multicolumn{2}{c|}{\% Overall} \\ \hline
Object     & method in\cite{Pas2015UsingGT} &Proposed &method in\cite{Pas2015UsingGT} &Proposed \\ \hline
Individual    &73      & 85        & 88   & 94     \\ \hline
clutter       &51      & 82        &73     & 93     \\ \hline
\end{tabular}
\end{table}

Another comparison with the method of paper \cite{Pas2015UsingGT} is given in table \ref{tab:indobjcomp} where the performance is calculated based on generated hypothesis. The comparison of overall performance also give in \ref{tab:indobjcomp}. The reasons for improvement at hypothesis generation stage has already been discussed. In \cite{Pas2015UsingGT} authors have enhanced the performance by training a SVM classifier to detect valid grasp out of a number of hypotheses created using HoG features. HoG feature depends on image gradient and, if there is no change of color in the boundary region then this feature may not be very useful. Comparatively, our approach does not require any training phase and can be applied in real time also. We have chosen the testing objects of different shape and size and our algorithm detects reliable grasping position for each object.  Some examples of detected handles for single objects is shown in figure[5]. For each object two image are given, one for 3D grasping pose, location and other one is a pair of boundary lines.

\begin{figure}[t]
\centerline{
\includegraphics[width=2cm,height=2cm]{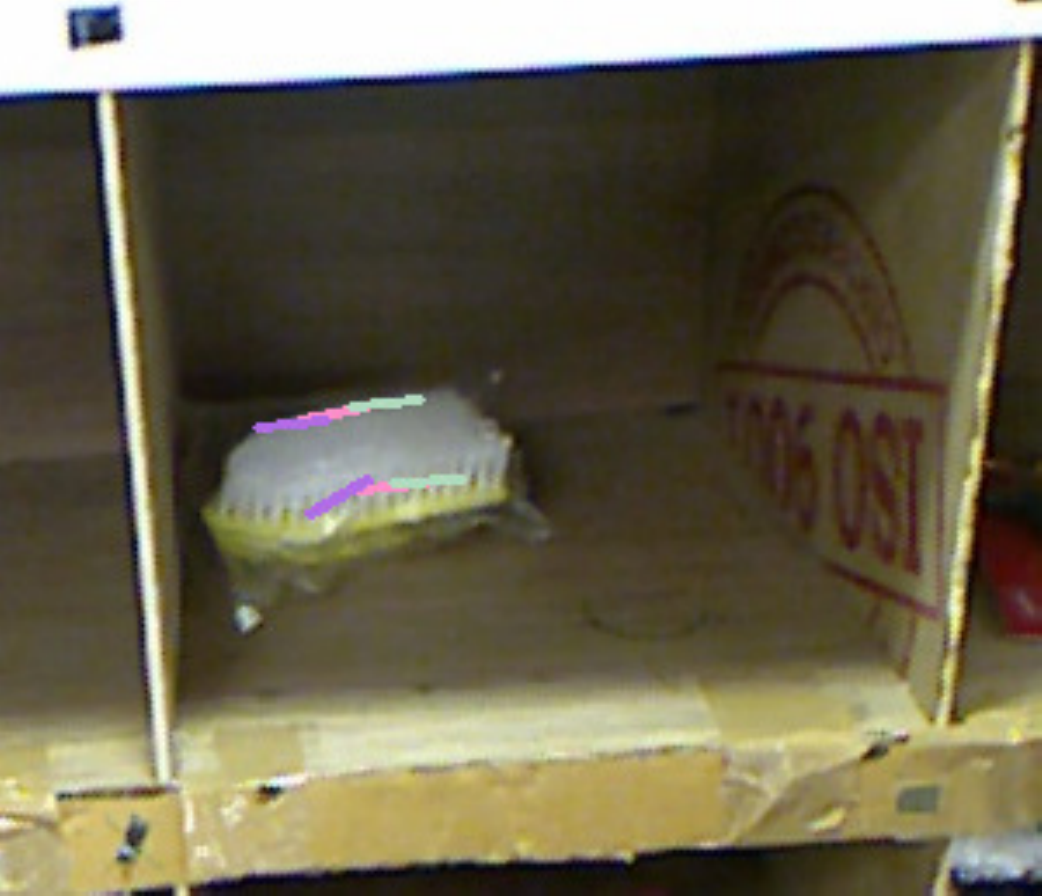}
\includegraphics[width=2cm,height=2cm]{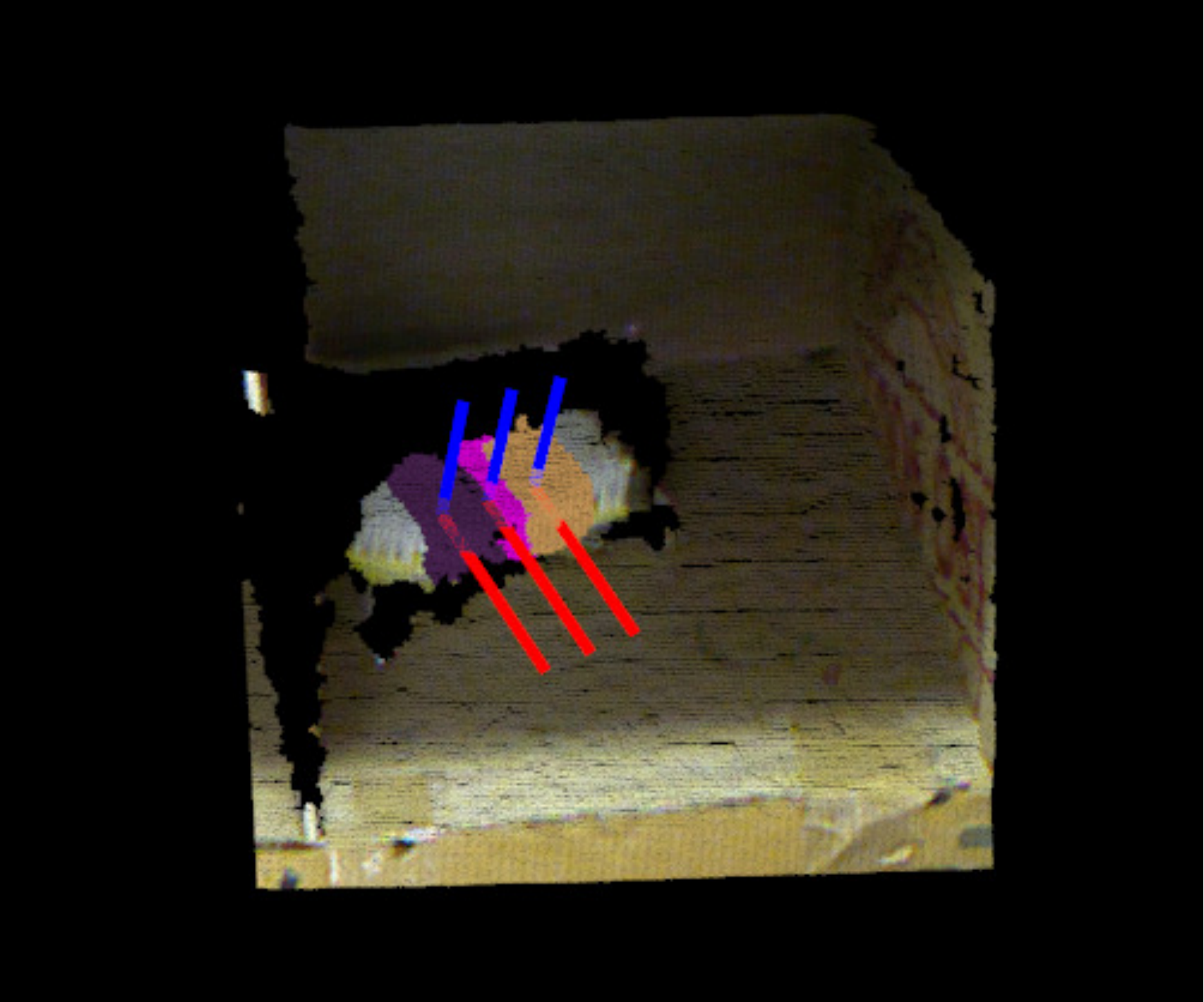}
\includegraphics[width=2cm,height=2cm]{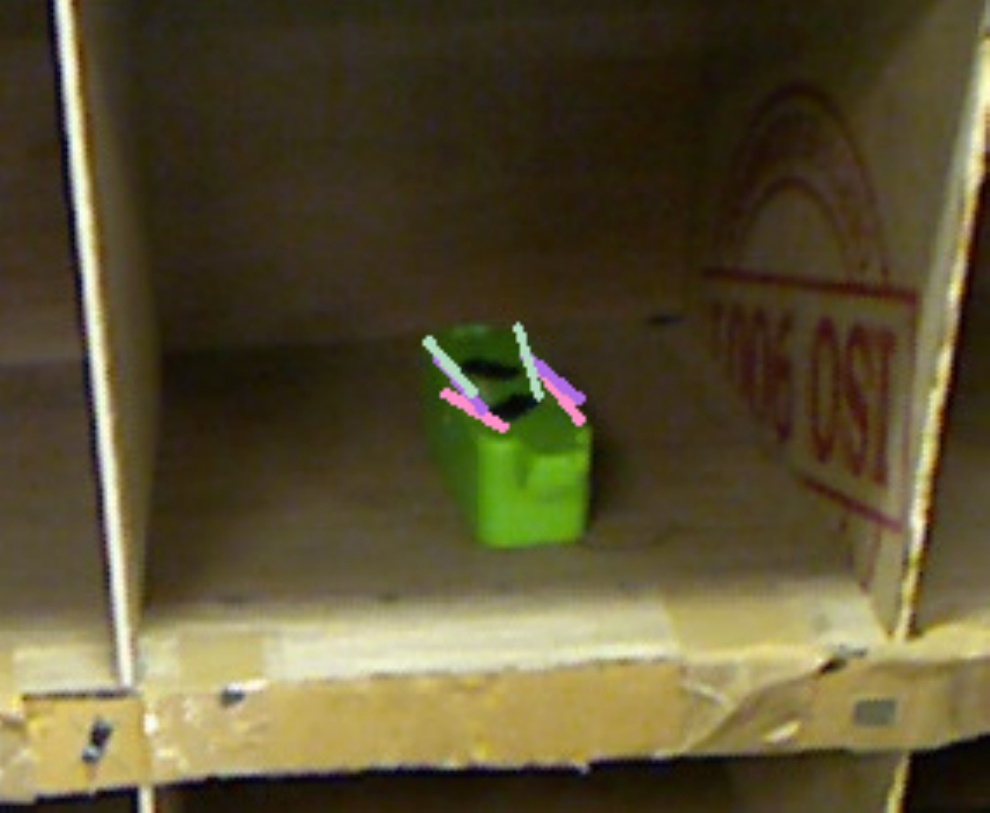}
\includegraphics[width=2cm,height=2cm]{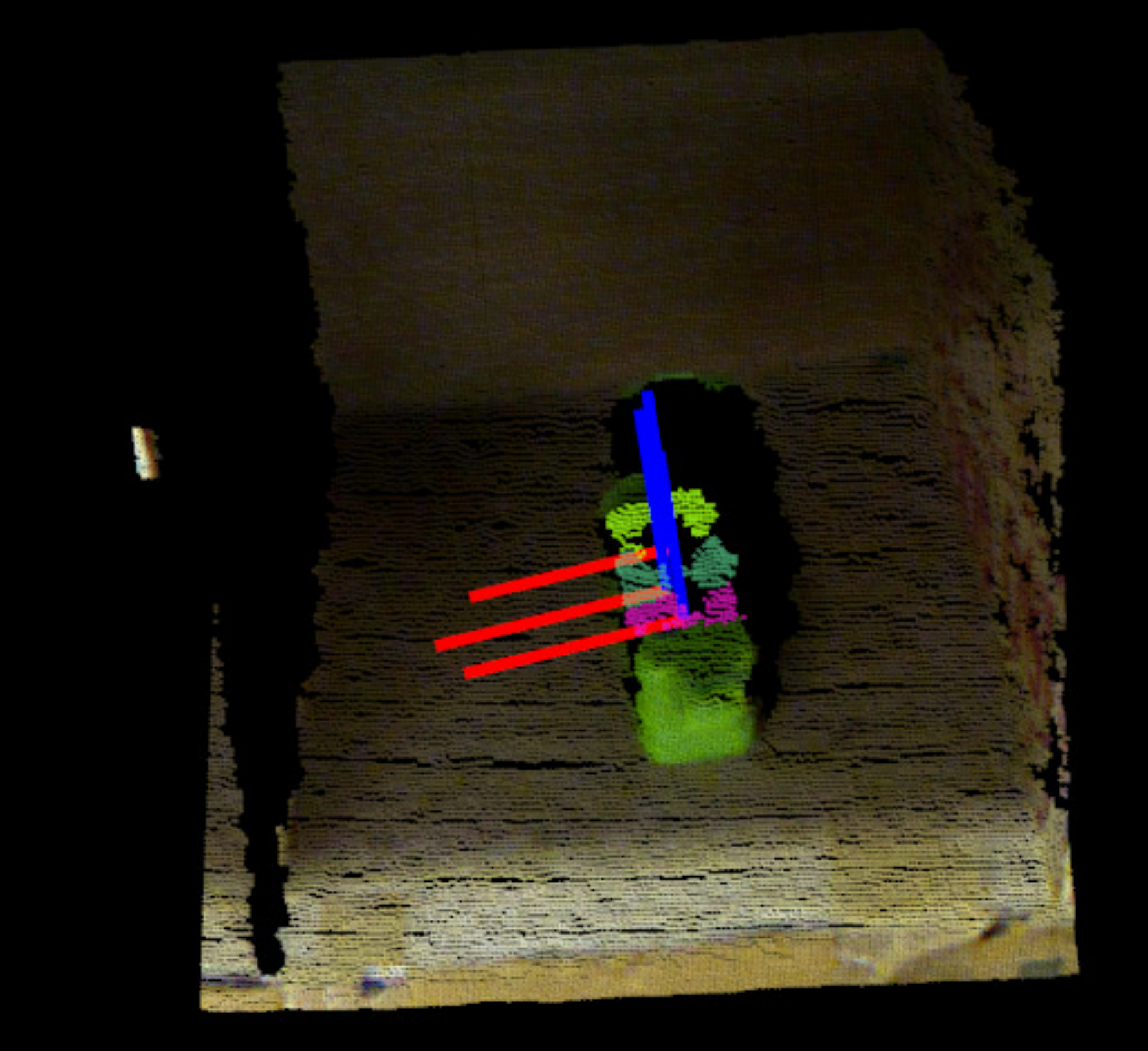}
\includegraphics[width=2cm,height=2cm]{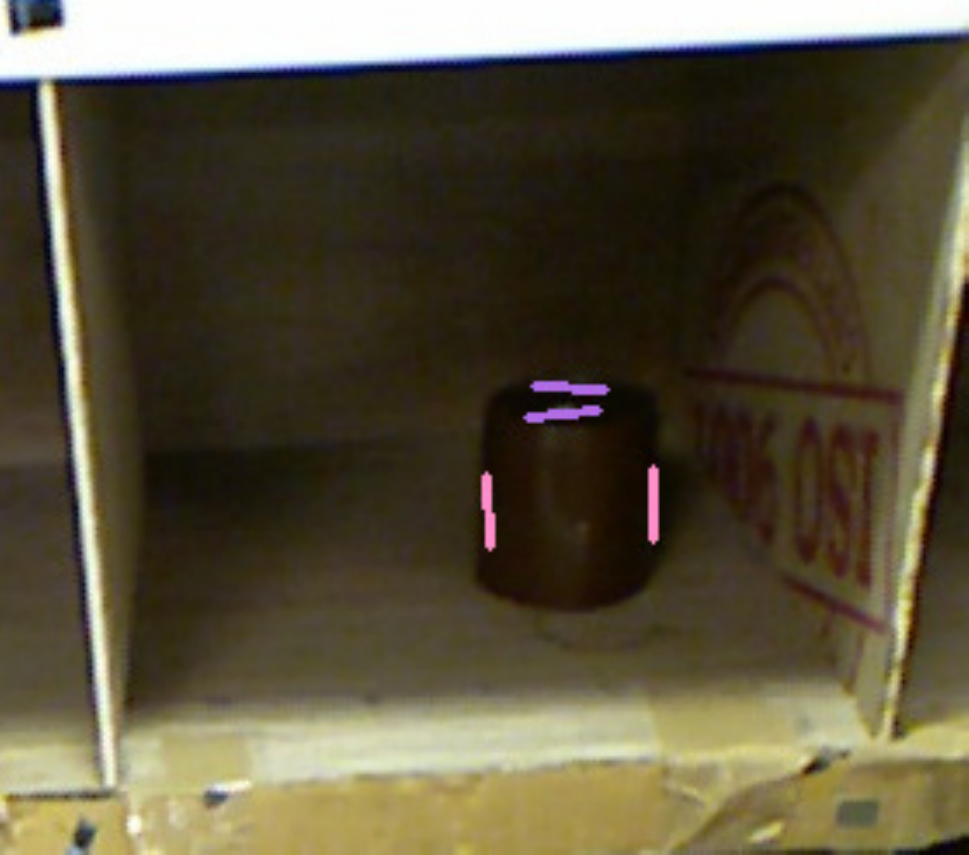}
\includegraphics[width=2cm,height=2cm]{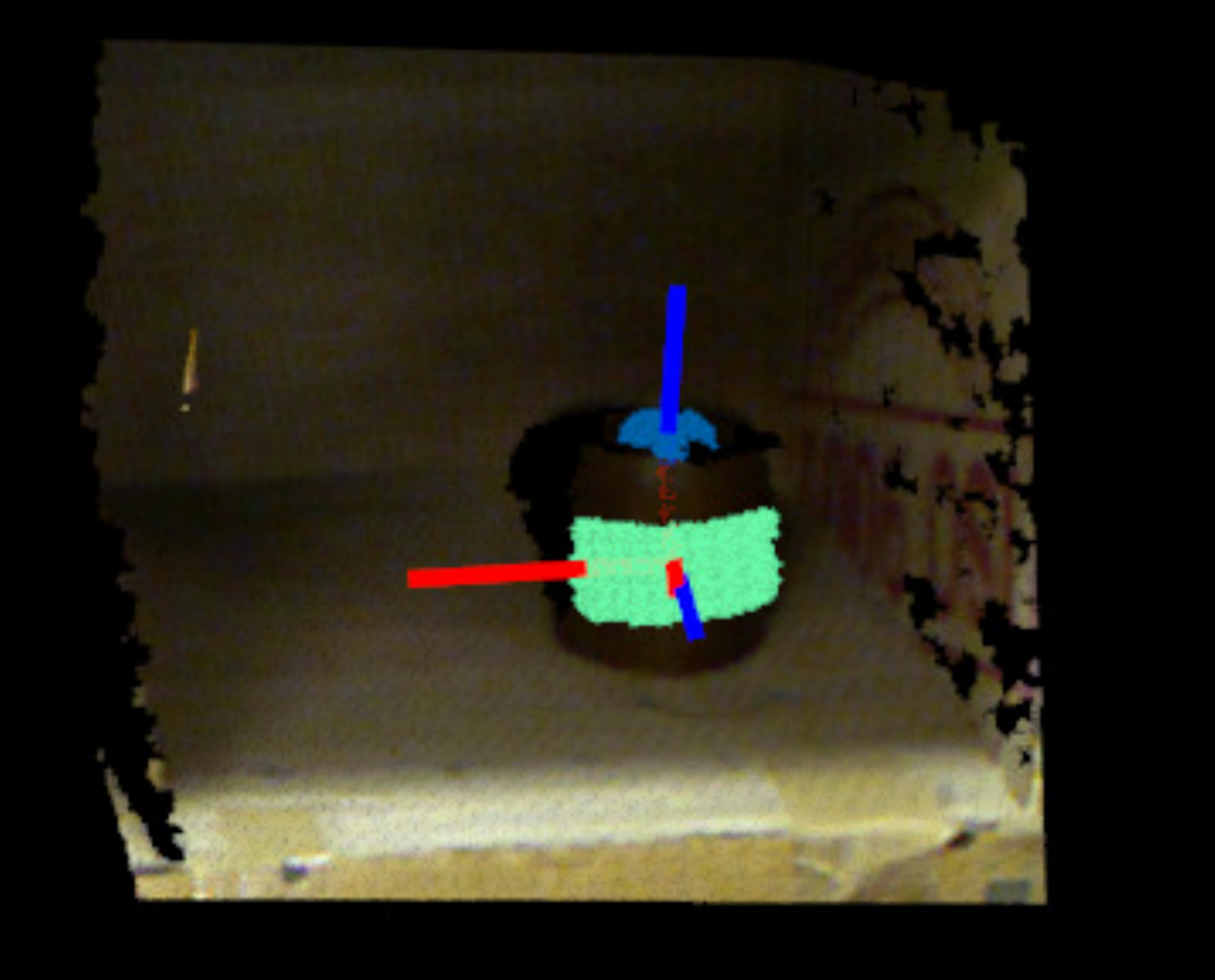}
}
\centerline{
\includegraphics[width=2cm,height=2cm]{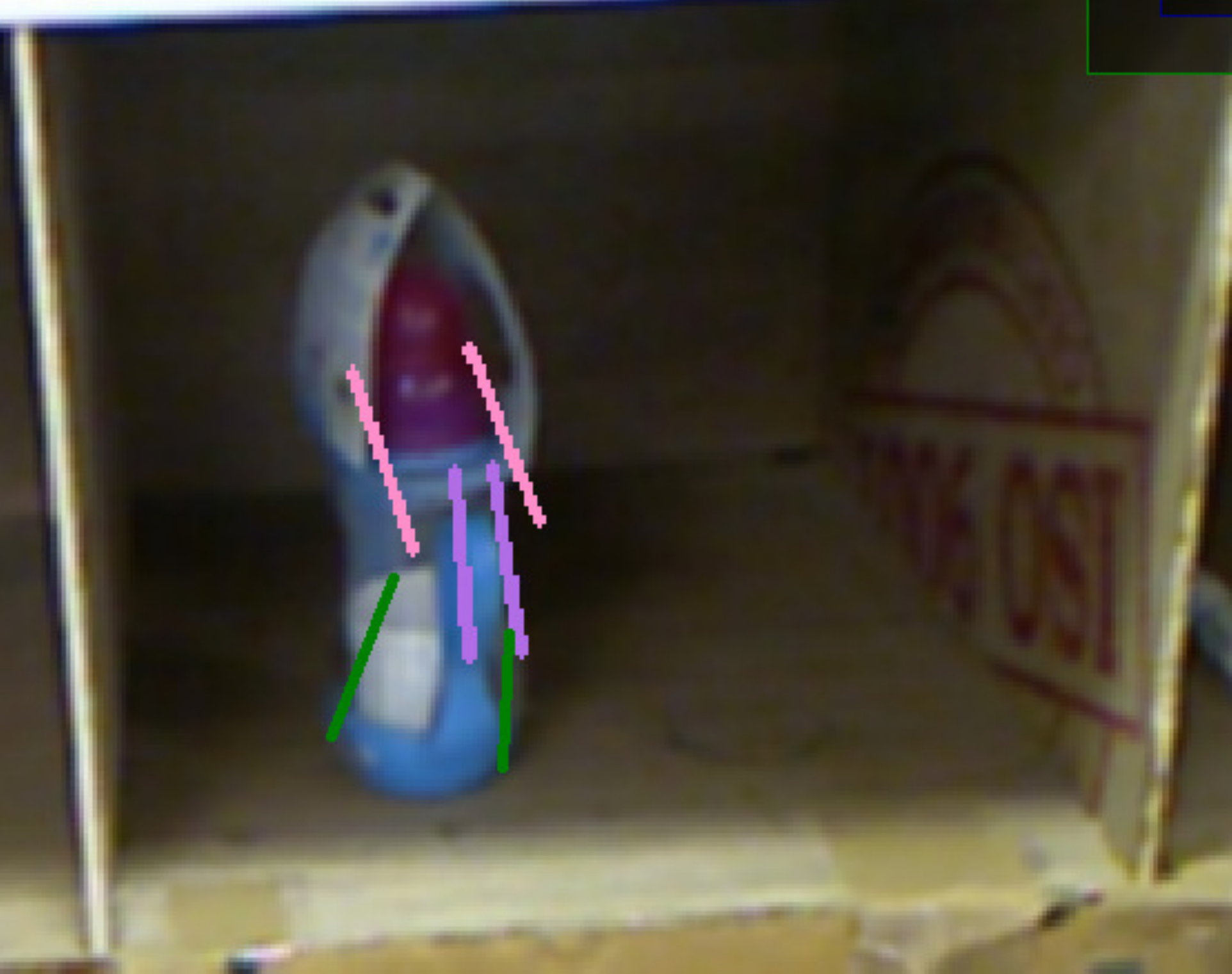}
\includegraphics[width=2cm,height=2cm]{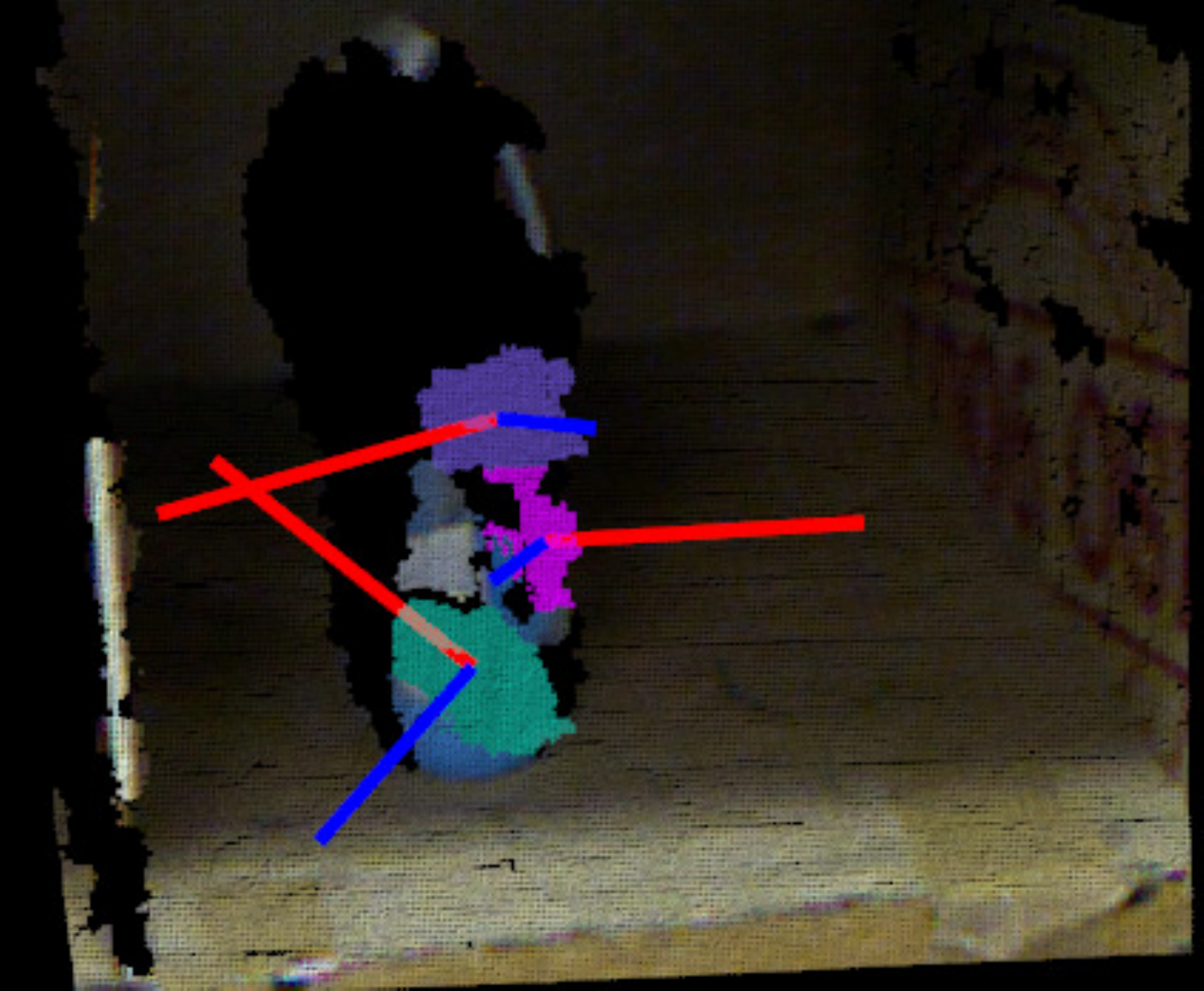}
\includegraphics[width=2cm,height=2cm]{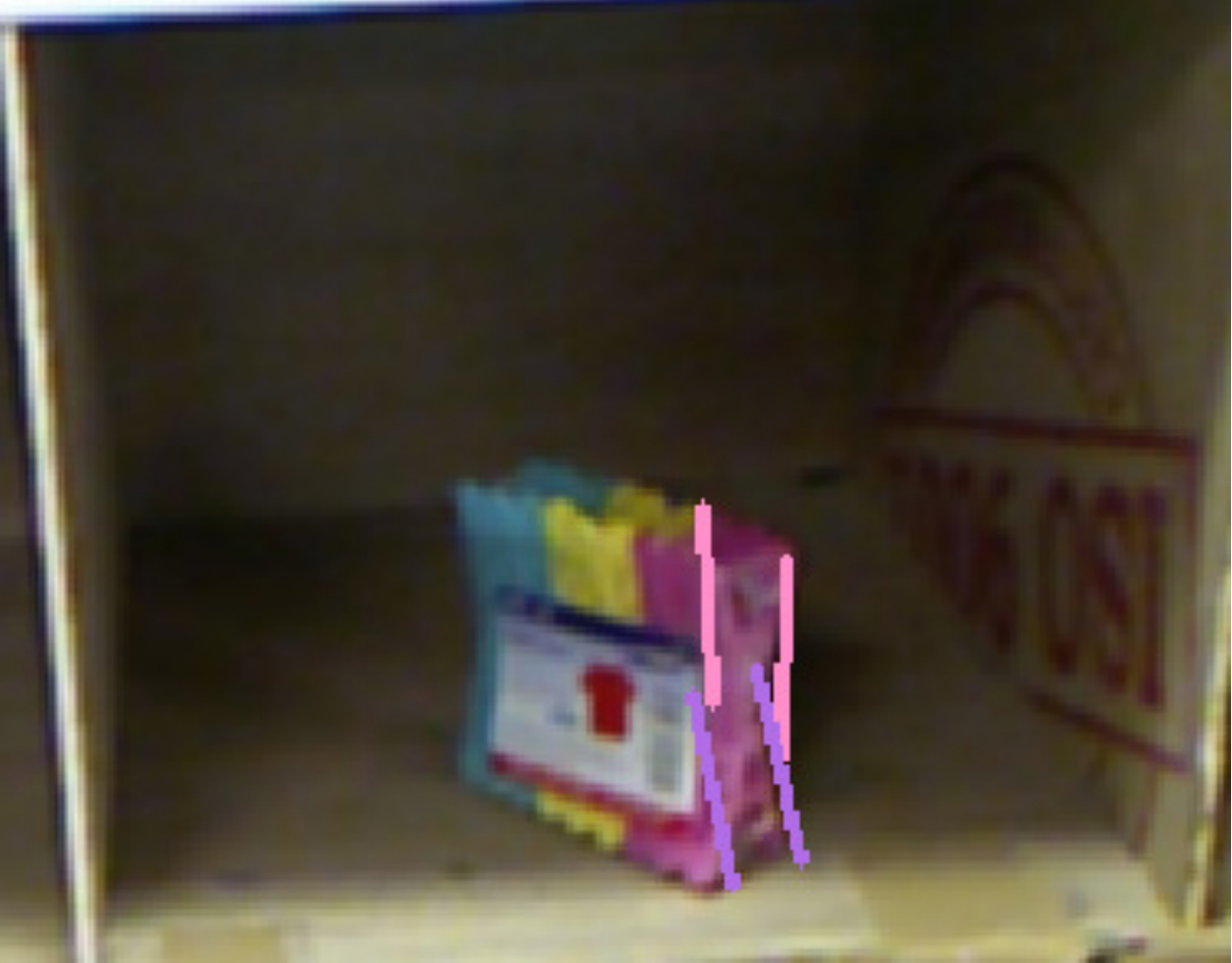}
\includegraphics[width=2cm,height=2cm]{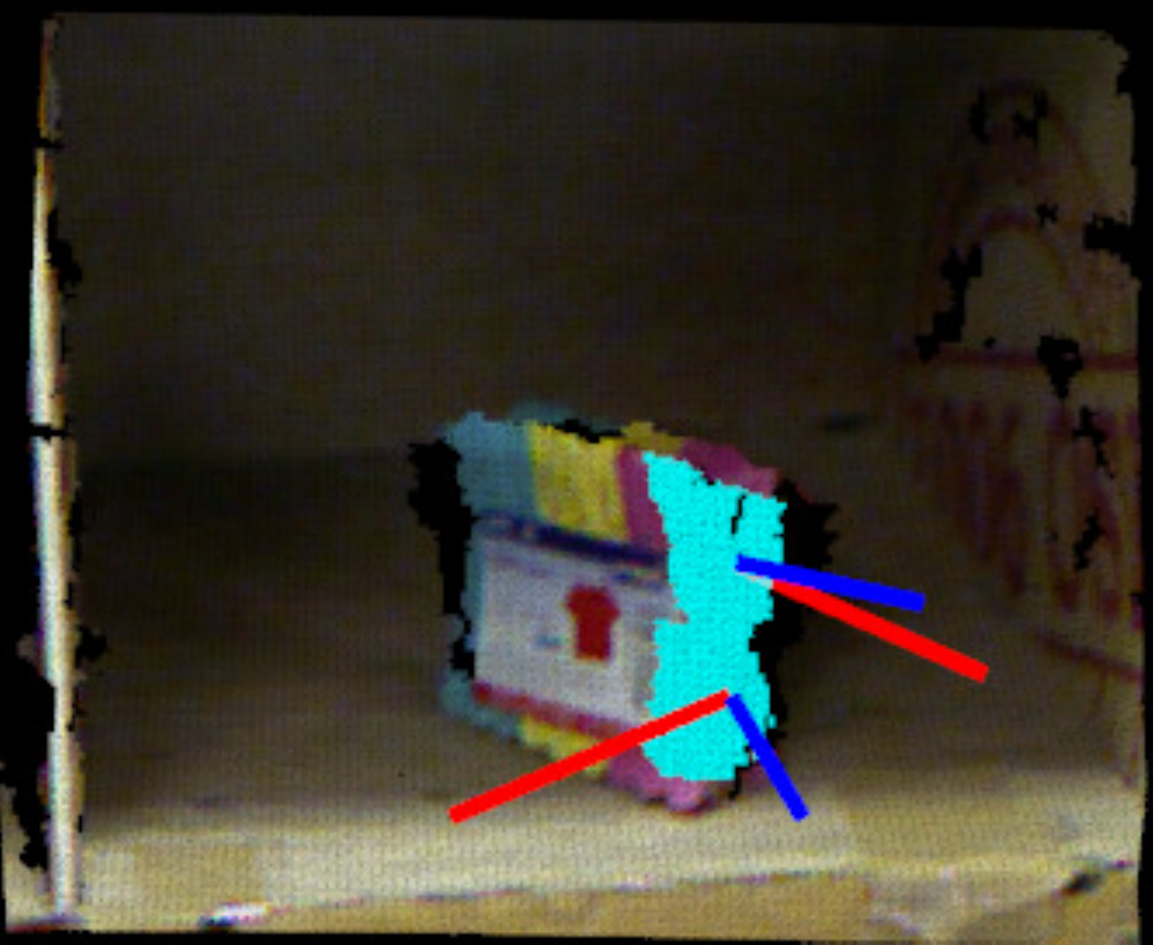}
\includegraphics[width=2cm,height=2cm]{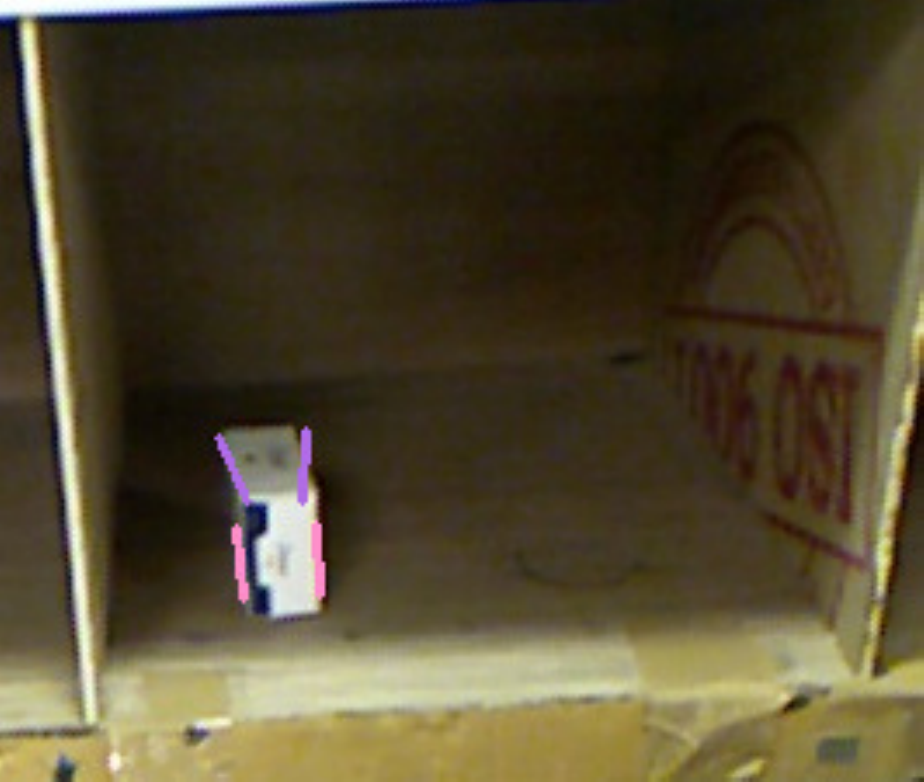}
\includegraphics[width=2cm,height=2cm]{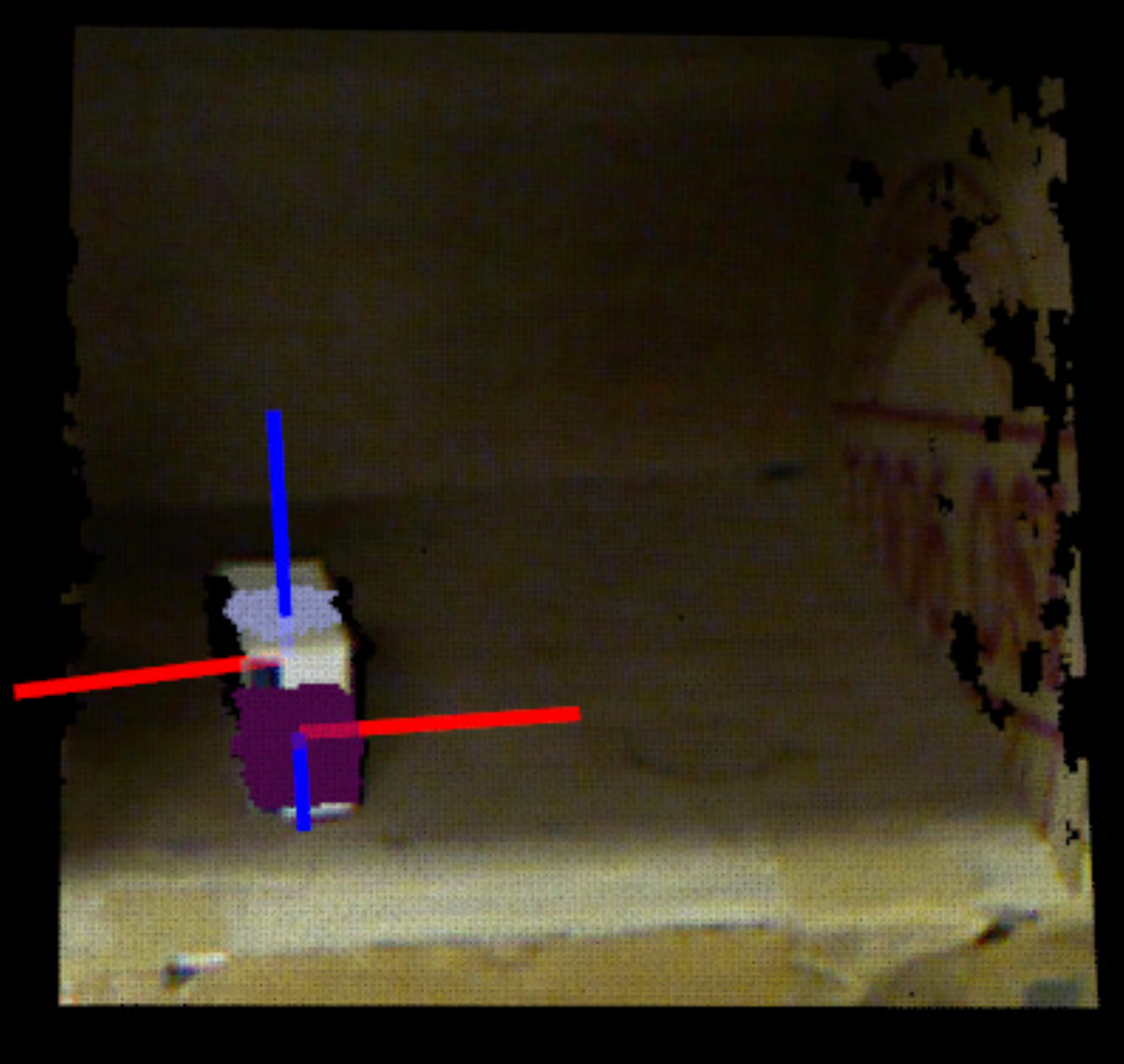}
}
\label{fig:singleObj}
\caption{Grasping handles in single object scene.}
\end{figure}

Most of the real life application of grasping include an unpredictable environment with noisy 3D input data. Only for a few cases there are single objects or multiple objects in a frame with sufficient gap between each object. Any grasping algorithm has to be effective enough to be applied into densely cluttered scene. In this work, our main target is to pick from cluttered scene and we have extensively evaluated the pipeline with such type of workspace. The objects in the frame may exhibit partial or full occlusion. Example given in Figure[6] show the performance of our method in an extremely cluttered workspace. With 85\% precision the proposed method generates grasp hypothesis and the overall performance with cluttered scene is 92\%. This data is compared with the work described in \cite{Pas2015UsingGT} in table \ref{tab:indobjcomp} where we can see huge improvements in results for grasp hypothesis generation and overall performance. In table \ref{tab:padcomp} the performance comparison is given with various publicly available datasets. Although lot of datasets are available for object detection/segmentation, only a few can be used for grasping. Again the form of the input for different method varies widely, making some of the datasets unsuitable to our method. We have given the performance comparison on a few grasping datasets and on a few datasets from segmentation/reconstruction as format of such datasets matches our requirement. Cornell grasping dataset \cite{lenz2015deepgrasp} is a famous dataset for grasping where their work is focused on grasping with deep learning methods. The accuracy of their method is 93.7\% and we get 94\% precision. However this dataset has only one object per frame but the objects do vary in size and shape. The ECCV dataset \cite{Aldoma2012}, Kinect Dataset \cite{SegIROS11} and the Willowgarage dataset \cite{willow} have multiple object per frame but the objects are placed away from each other. These dataset are created for segmentation purpose, hence do not consider challenging cases for grasping and they don't have corresponding annotations either. Due to unavailability of dataset for which the extent of clutter can be comparable to practical workspace, we have tested the proposed method with our own dataset. The precision is  93\% as shown in table \ref{tab:padcomp}.

\begin{figure}
\centerline{
\includegraphics[width=2.5cm, height=2cm]{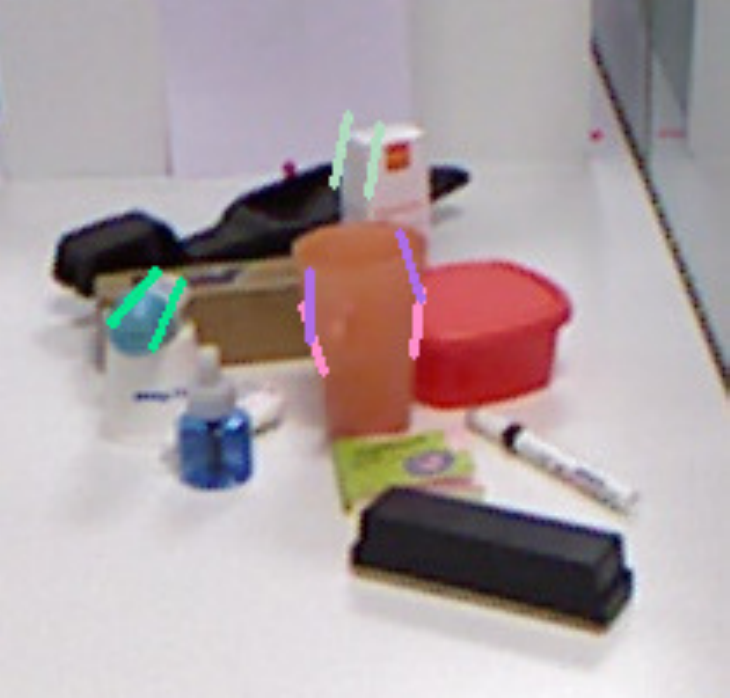}
\includegraphics[width=2.5cm, height=2cm]{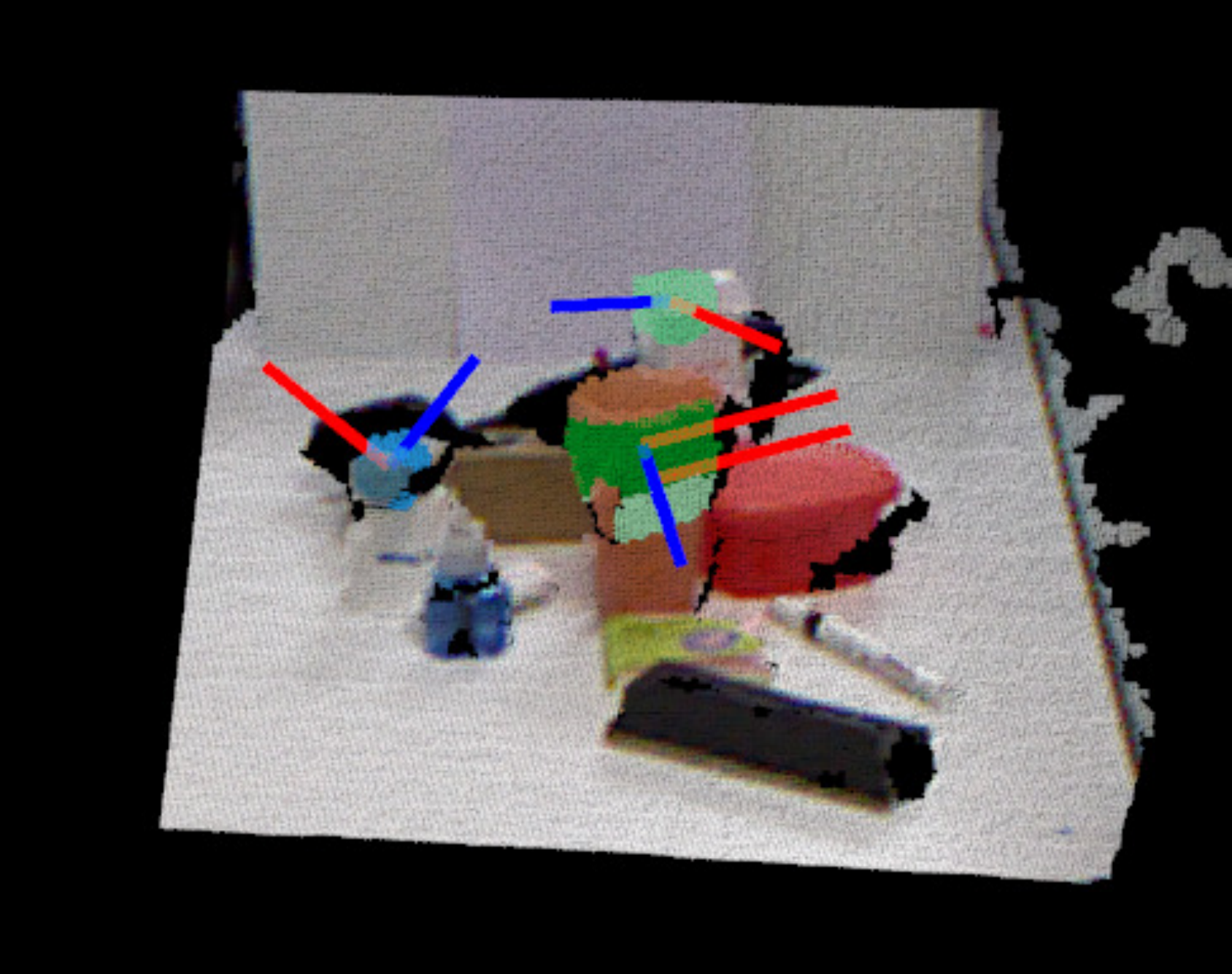}
\includegraphics[width=2.5cm, height=2cm]{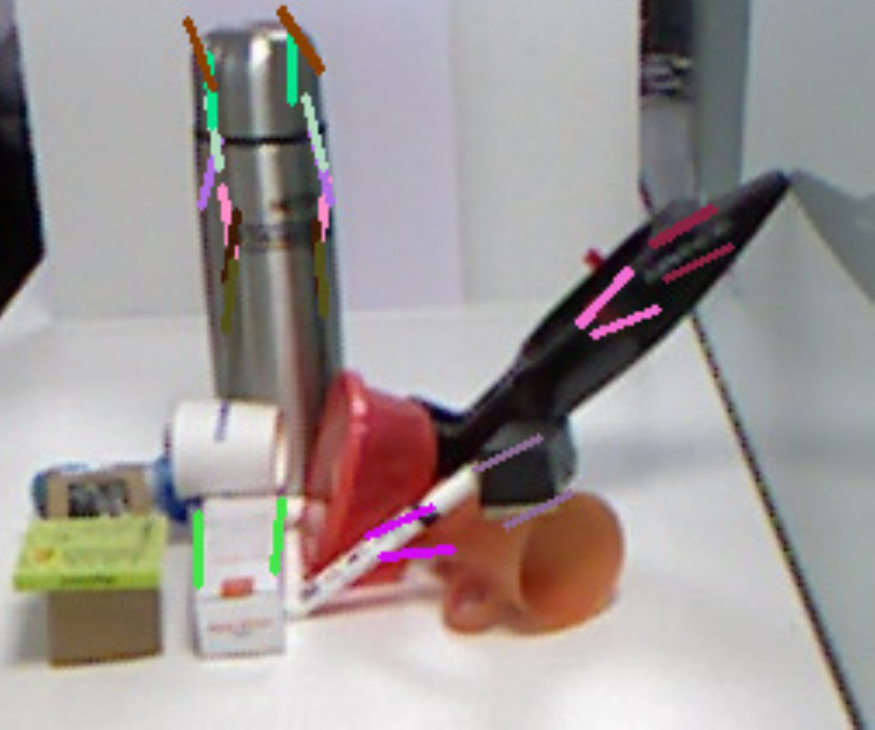}
\includegraphics[width=2.5cm, height=2cm]{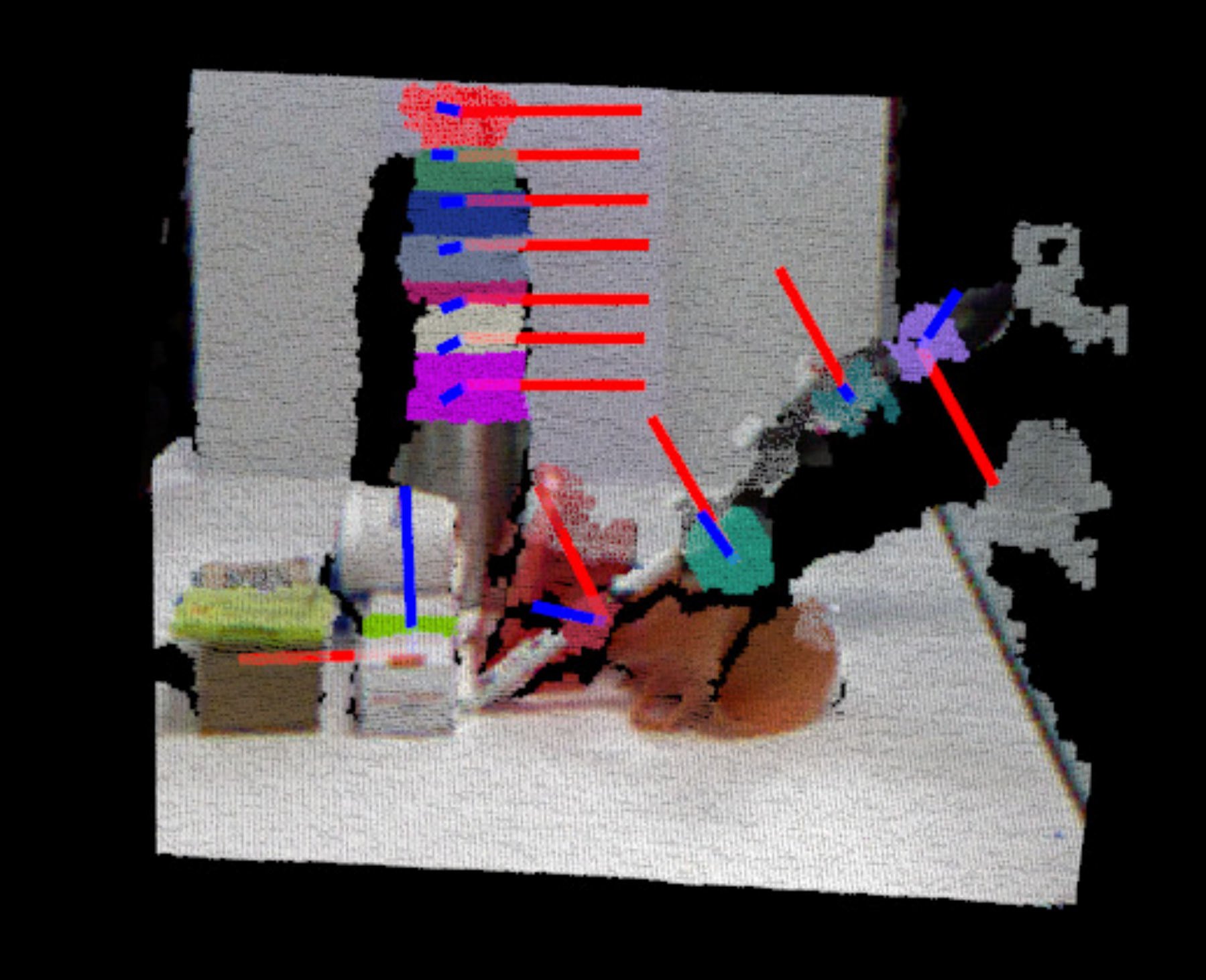}
}
\centerline{
\includegraphics[width=2.5cm, height=2cm]{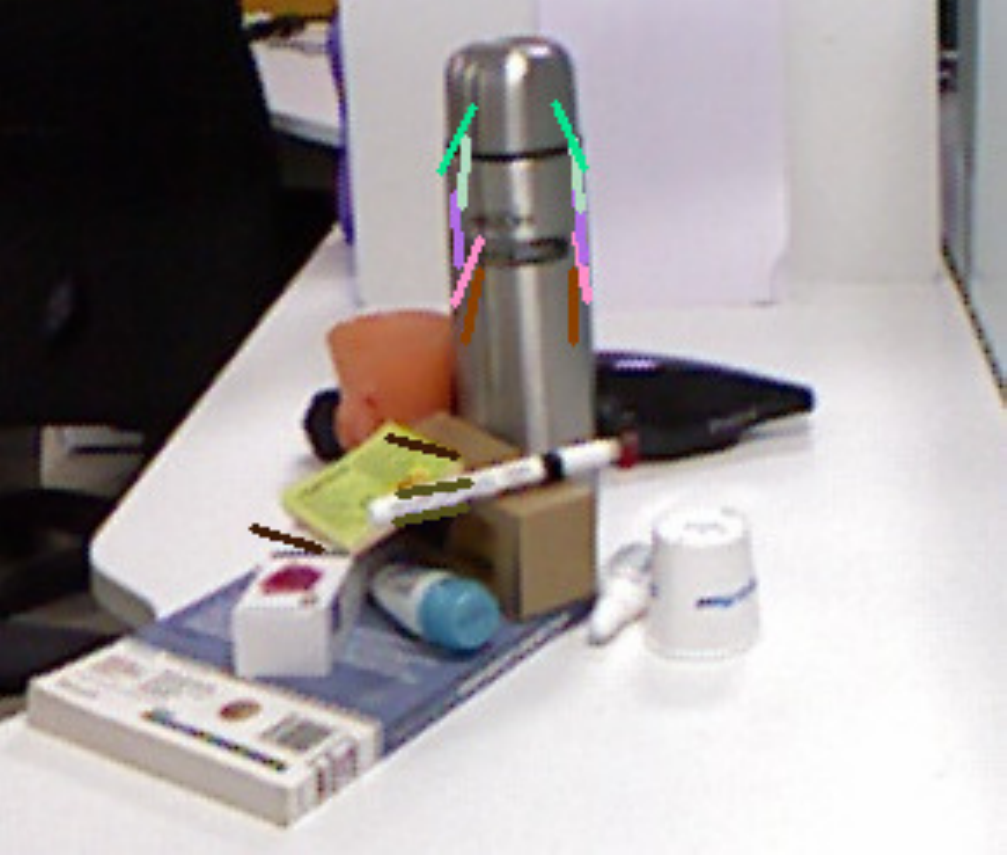}
\includegraphics[width=2.5cm, height=2cm]{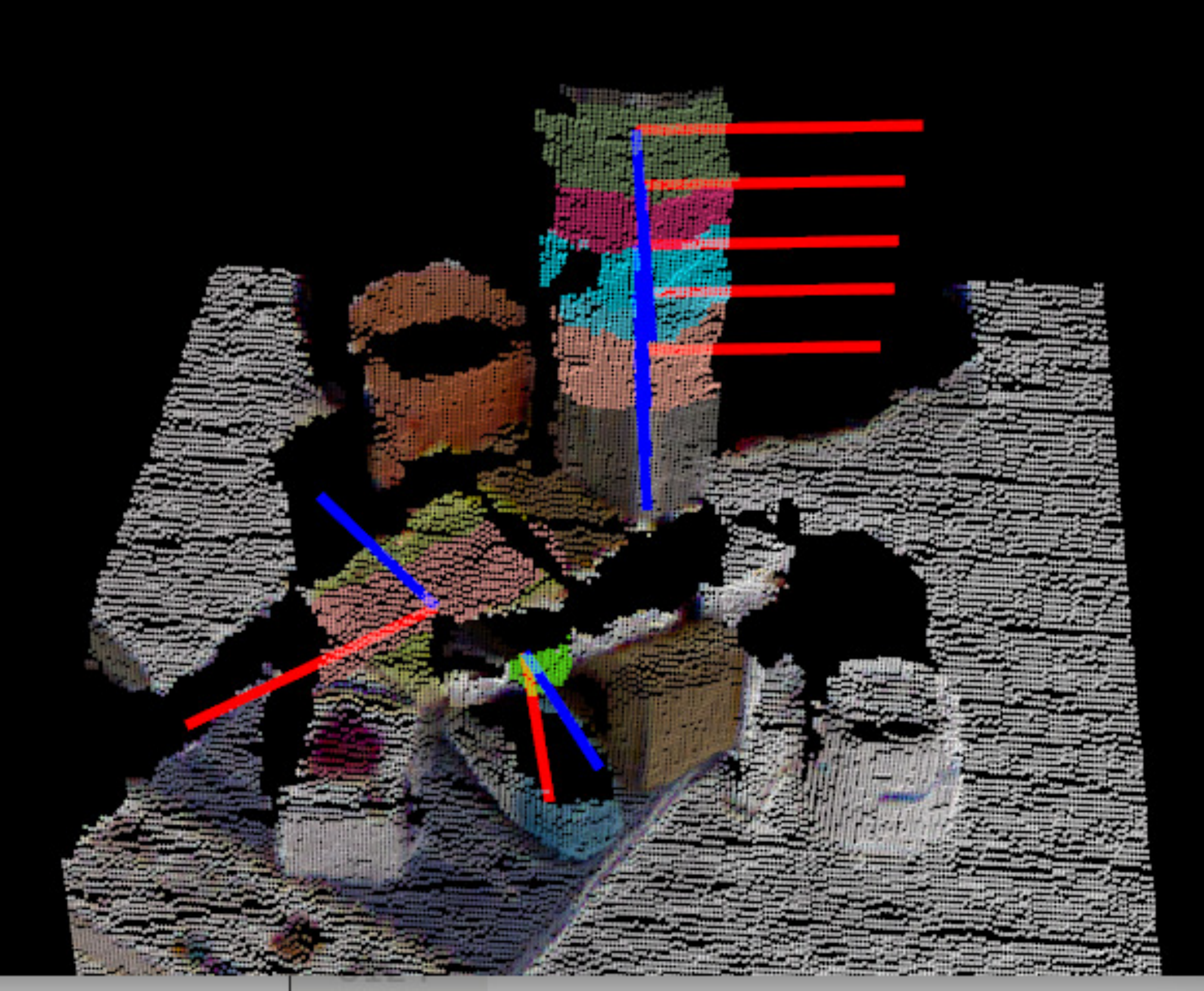}
\includegraphics[width=2.5cm, height=2cm]{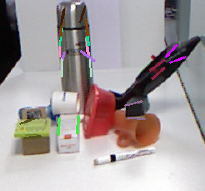}
\includegraphics[width=2.5cm, height=2cm]{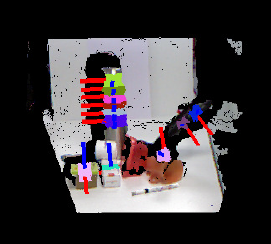}
}
\centerline{
\includegraphics[width=2.5cm, height=2cm]{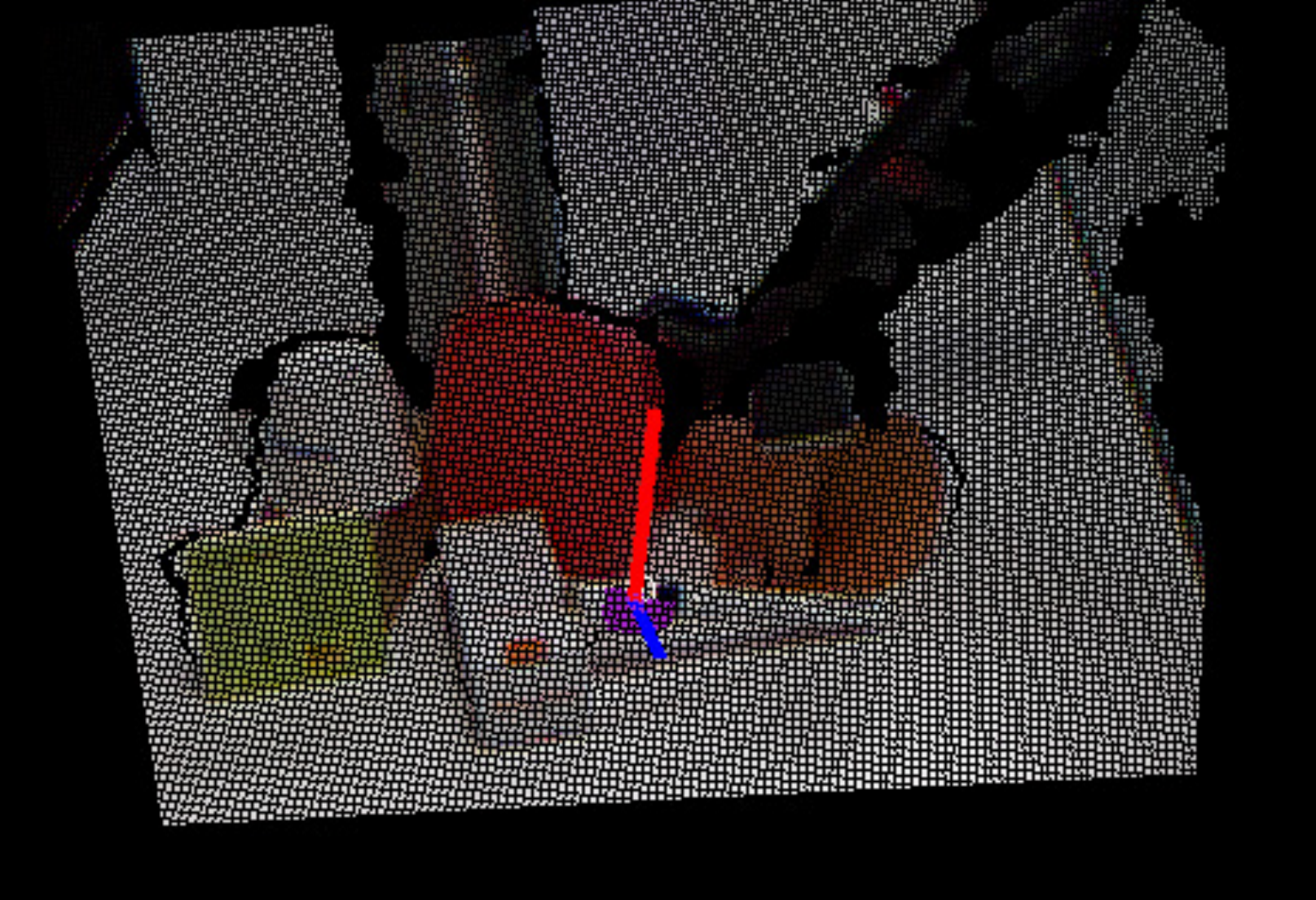}
\includegraphics[width=2.5cm, height=2cm]{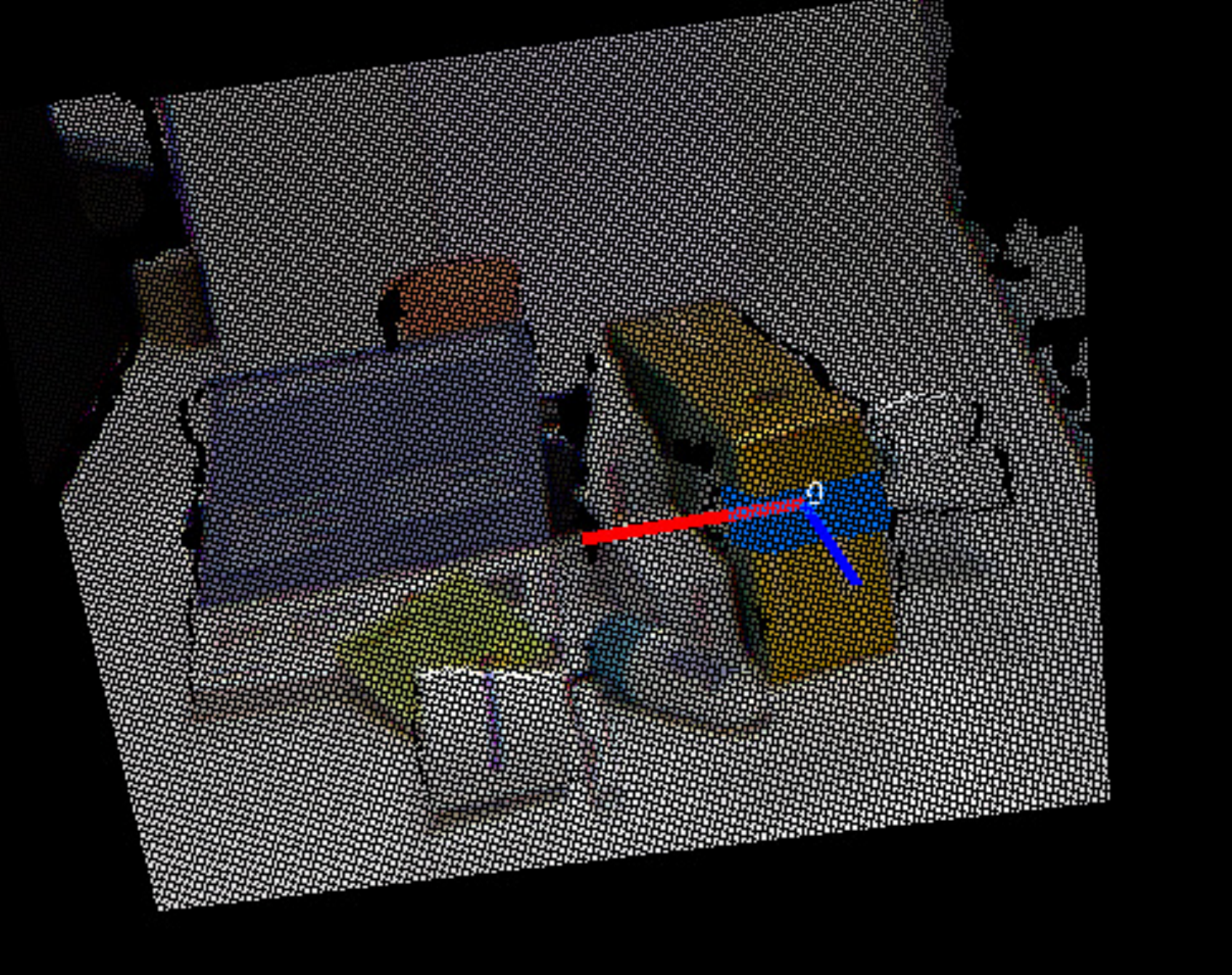}
\includegraphics[width=2.5cm, height=2cm]{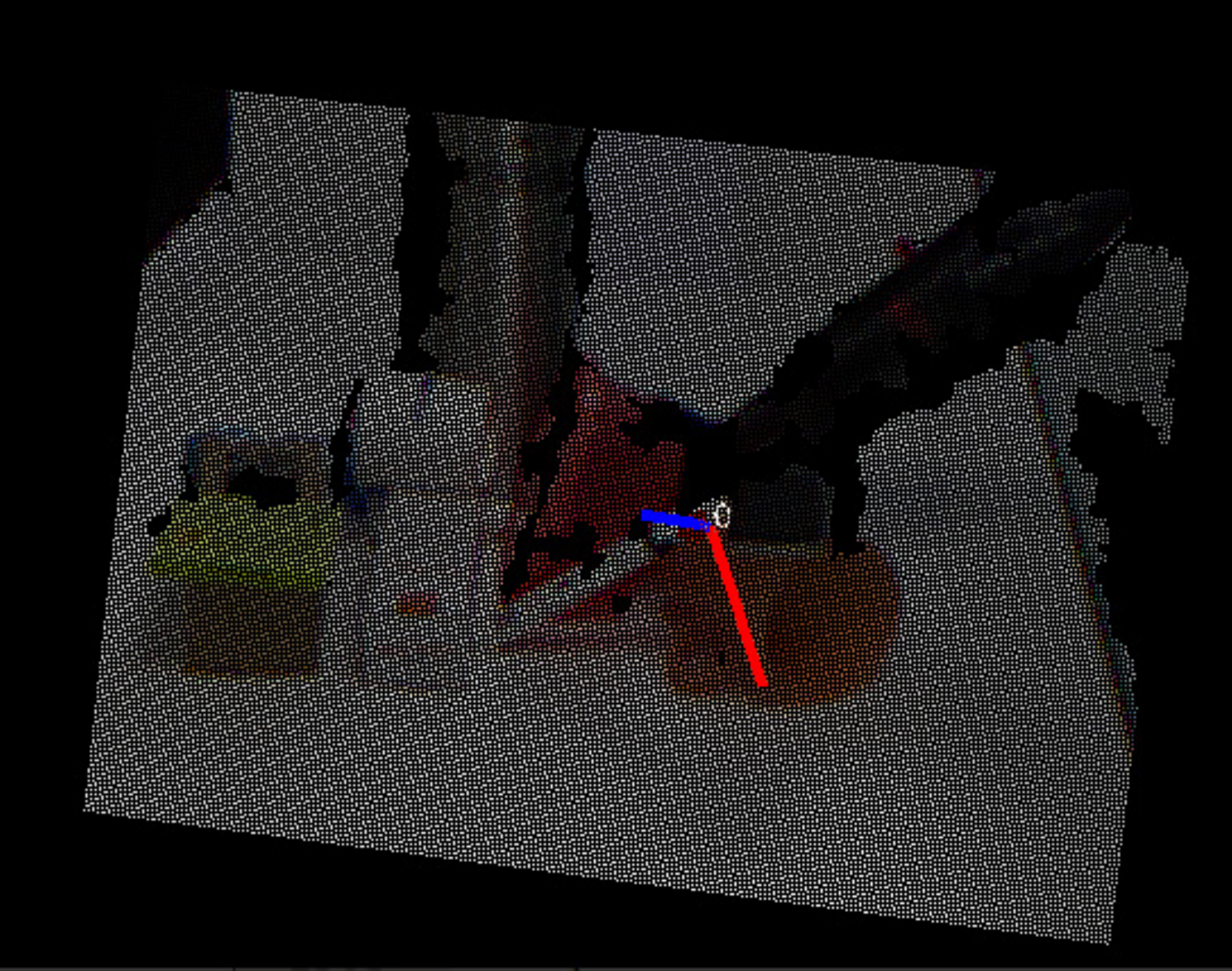}
\includegraphics[width=2.5cm, height=2cm]{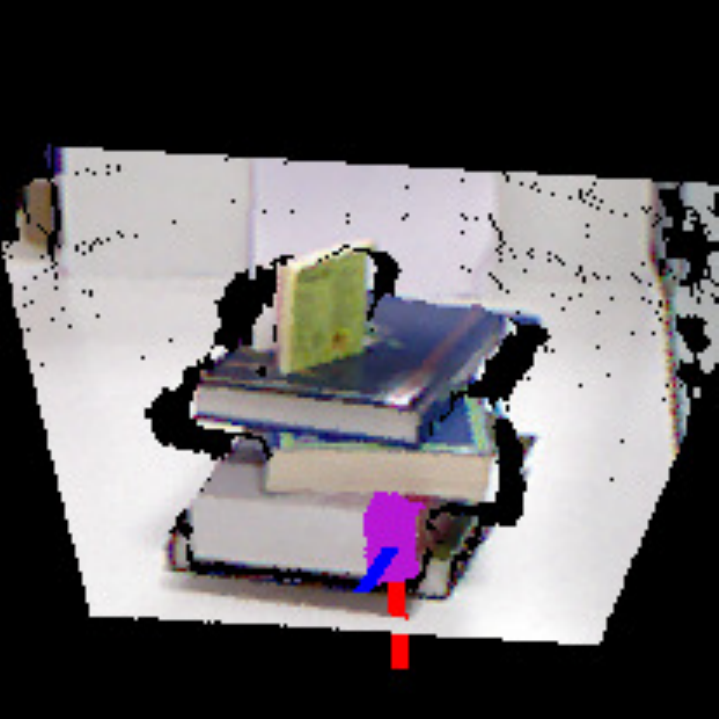}
\includegraphics[width=2.5cm, height=2cm]{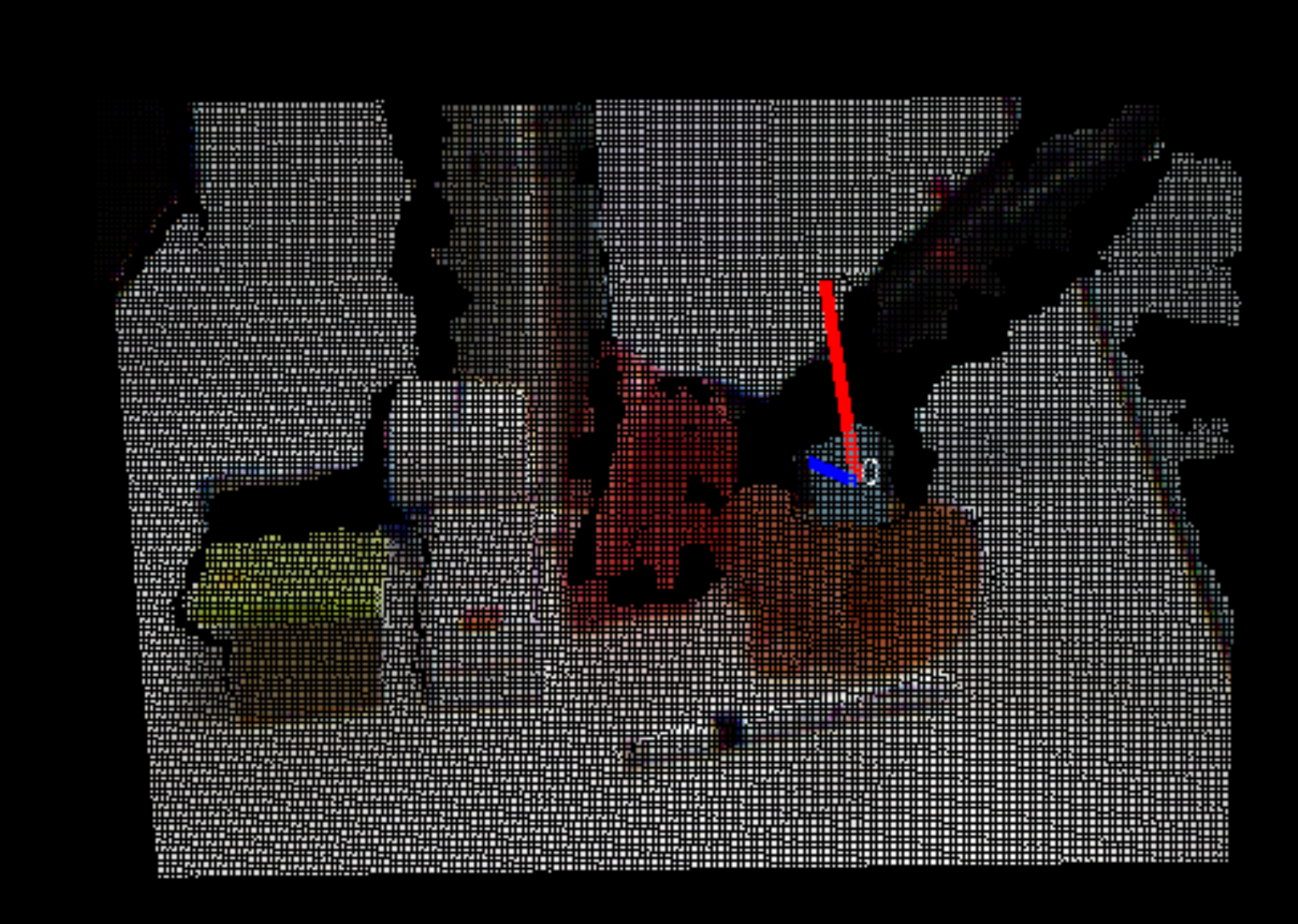}
}

\label{fig:clutter}
\caption{First two row shows grasping handles on densely cluttered scene and last one is best handle returned by the ranking algorithm }
\end{figure}

\begin{table}[!t]
\centering
\caption{Performance Comparison on Various Publicly Available Datasets}
\label{tab:padcomp}
\small
\begin{tabular}{|>{\centering\arraybackslash}m{0.3cm}|c|>{\centering\arraybackslash}m{1.5cm}|>{\centering\arraybackslash}m{1.5cm}|} \hline
  &    & \multicolumn{2}{c|}{\% Precision} \\ \hline
S. No. & \textbf{Dataset} & Proposed Method & Other Methods \cite{Pas2015UsingGT}  \\ \hline
2 & Cornell Dataset \cite{lenz2015deepgrasp}   & 95\%         & 93.7\% \cite{lenz2015deepgrasp}     \\ \hline
3 & ECCV \cite{Aldoma2012}           & 91\%    & 63\%     \\ \hline
4 & Kinect Dataset  \cite{SegIROS11} & 92\%    &62\%      \\ \hline
5 & Willow Garage \cite{willow}   & 96\%         & 70\%     \\ \hline
6 & Individual   & 94\%         & 88\%   \\ \hline
7 & Cluttered    & 93\%         & 64\%   \\ \hline
\end{tabular}
\end{table}

After getting all possible handle, we have to find the one using which robot will perform the act. Instead of choosing the one at random from the set of valid handles, we determine the best handle in the set by assigning score to each of them. Some example are shown in figure [6] 
Precision along with the effective selection of handles defines the success rate of our method which is around 95\%. Our algorithms run on real time and the average processing time for a complete frame with around 50K data point is approximately 500 ms to 1 second on a Linux laptop with a i7 processor and 16 GB RAM.

\section{Conclusions}\label{sec:conc}
This paper proposes a novel technique for findings grasping handlers of objects placed in a cluttered environment. By merging color based edge and depth edge we detected a pair of boundary lines for each potential handle. These potential handlers are then verified as valid handle only if the angle between the two boundary lines is less than a threshold and if there is no obstacle on the way for that handle. Finally, from a bunch of valid handles, we choose the best handle based on the minimum cost using a cost function. Our proposed method, when tested on real-life datasets, outperforms state of the art methods by a huge margin.

\bibliography{egbib} 
\bibliographystyle{ieeetr}

\end{document}